\documentclass[10pt,twocolumn,letterpaper]{article}

\usepackage[pagenumbers]{cvpr} 

\usepackage{graphicx}
\usepackage{amsmath}
\usepackage{amssymb}
\usepackage{pifont} 
\usepackage{booktabs}
\usepackage{enumitem}
\usepackage[dvipsnames,svgnames,x11names]{xcolor}

\definecolor{citecolor}{HTML}{0071bc}
\usepackage[pagebackref=true,breaklinks=true,colorlinks,citecolor=citecolor,bookmarks=false]{hyperref}

\usepackage[capitalize]{cleveref}
\crefname{section}{Sec.}{Secs.}
\Crefname{section}{Section}{Sections}
\Crefname{table}{Table}{Tables}
\crefname{table}{Tab.}{Tabs.}

\def\cvprPaperID{xxxx}
\def\confName{CVPR}
\def\confYear{2022}
\newcommand{\cmark}{\ding{51}}
\newcommand{\xmark}{\ding{55}}
\newlength\savewidth\newcommand\shline{\noalign{\global\savewidth\arrayrulewidth
  \global\arrayrulewidth 1pt}\hline\noalign{\global\arrayrulewidth\savewidth}}
\newcommand{\tablestyle}[2]{\setlength{\tabcolsep}{#1}\renewcommand{\arraystretch}{#2}\centering\small}
\makeatletter\renewcommand\paragraph{\@startsection{paragraph}{4}{\z@}
  {.5em \@plus1ex \@minus.2ex}{-.5em}{\normalfont\normalsize\bfseries}}\makeatother

\begin{document}

\title{GroupViT: Semantic Segmentation Emerges from Text Supervision}

\author{
Jiarui Xu\textsuperscript{1}\thanks{Jiarui Xu was an intern at NVIDIA during the project.} \hspace{2mm}
Shalini De Mello\textsuperscript{2} \hspace{2mm}
Sifei Liu\textsuperscript{2} \hspace{2mm}
Wonmin Byeon\textsuperscript{2} 
\\
Thomas Breuel\textsuperscript{2} \hspace{2mm}
Jan Kautz\textsuperscript{2} \hspace{2mm} 
Xiaolong Wang\textsuperscript{1}\\
\vspace{1mm}
\textsuperscript{1}UC San Diego \qquad \textsuperscript{2}NVIDIA \\ 
}
\maketitle

\vspace{-2.5em}
\begin{abstract}
    Grouping and recognition are important components of visual scene understanding, e.g., for object detection and semantic segmentation. With end-to-end deep learning systems, grouping of image regions usually happens implicitly via top-down supervision from pixel-level recognition labels. 
    Instead, in this paper, we propose to bring back the grouping mechanism into deep networks, which allows semantic segments to emerge automatically with only text supervision. We propose a hierarchical Grouping Vision Transformer (GroupViT), which goes beyond the regular grid structure representation and learns to group image regions into progressively larger arbitrary-shaped segments. We train GroupViT jointly with a text encoder on a large-scale image-text dataset via contrastive losses. With only text supervision and without any pixel-level annotations, GroupViT learns to group together semantic regions and successfully transfers to the task of semantic segmentation in a zero-shot manner, i.e., without any further fine-tuning. 
    It achieves a zero-shot accuracy of 52.3\% mIoU on the PASCAL VOC 2012 and 22.4\% mIoU on PASCAL Context datasets, and performs competitively to state-of-the-art transfer-learning methods requiring greater levels of supervision. We open-source our code at \href{https://github.com/NVlabs/GroupViT}{https://github.com/NVlabs/GroupViT}.
\end{abstract}
\vspace{-1.5em}

\section{Introduction}
\vspace{-0.5em}

\begin{figure}[t]
    \centering
    \includegraphics[width=.99\linewidth]{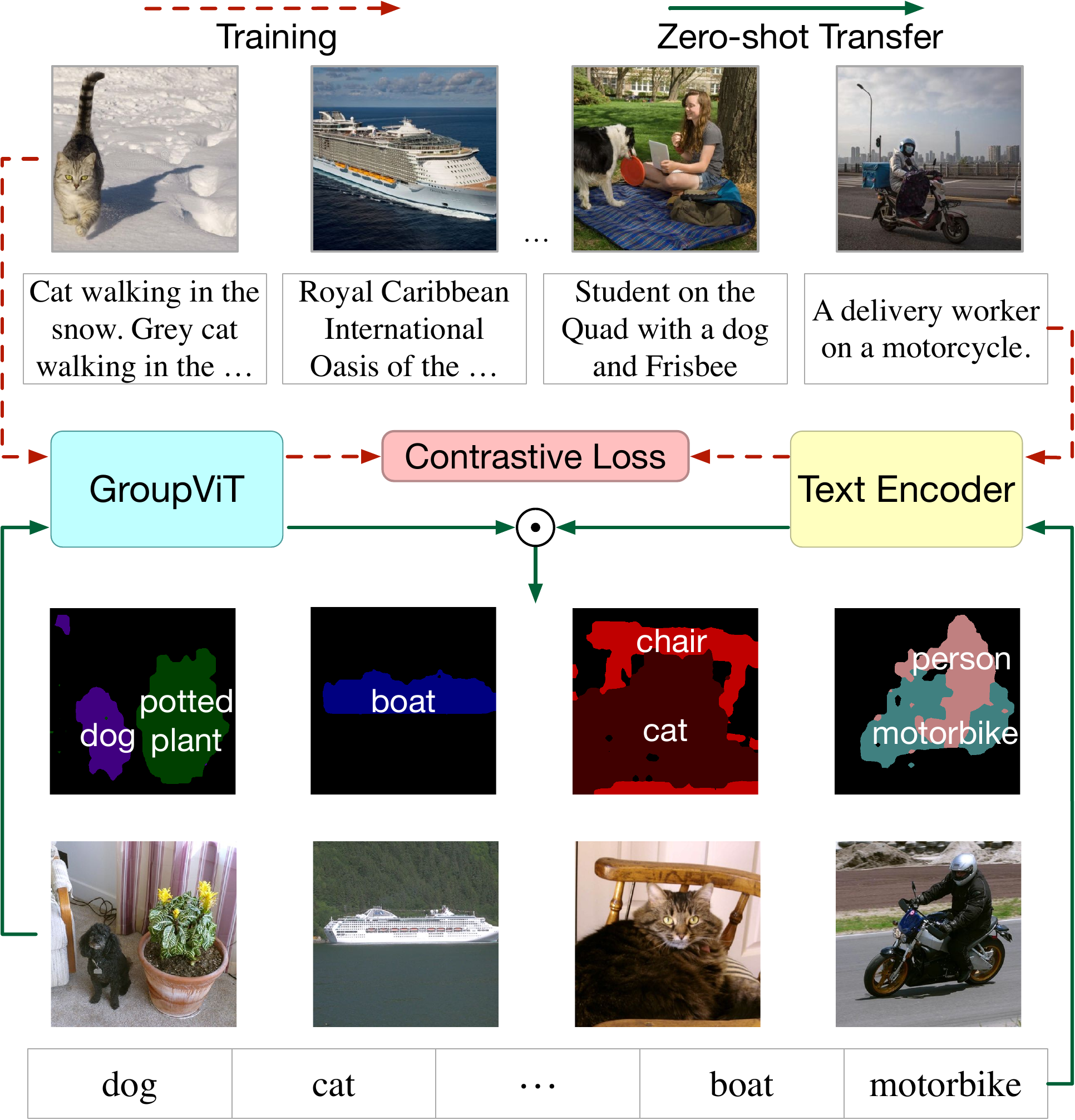}
    \vspace{.2em}
    \caption{
    \textbf{Problem Overview}. 
    First, we jointly train GroupViT and a text encoder using paired image-text data. With GroupViT, meaningful semantic grouping automatically emerges without any mask annotations. Then, we transfer the trained GroupViT model to the task of zero-shot semantic segmentation.}
    \label{fig:teaser}
    \vspace{-1.5em}
\end{figure}

Visual scenes are naturally composed of semantically-related groups of pixels. The relationship between grouping and recognition has been studied extensively in visual understanding even before the deep learning era~\cite{malik2001visual,malik2016three}. 
In bottom-up grouping, the idea is to first re-organize pixels into candidate groups and then to process each group with a recognition module. This pipeline has been successfully applied in image segmentation from superpixels~\cite{ren2003learning}, constructing region proposals for object detection~\cite{uijlings2013selective,zitnick2014edge} and semantic segmentation~\cite{arbelaez2012semantic}. Beyond bottom-up inference, top-down feedback from recognition can also provide signals to perform better visual grouping~\cite{tu2005image,zhu2007stochastic}.

However, on moving to the deep learning era, the ideas of explicit grouping and recognition have been much less separated and more tightly coupled 
in end-to-end training systems. Semantic segmentation, e.g., is commonly achieved via a Fully Convolutional Network~\cite{long2015fully}, where pixel grouping is only revealed at the output by recognizing each pixel's label. This approach eliminates the need to perform explicit grouping. While this method is very powerful and still delivers state-of-the-art performance, there are two major limitations that come with it: (i) learning is limited by the high cost of per-pixel human labels; and (ii) the learned model is restricted only to a few labeled categories and cannot generalize to unseen ones.

Recent developments in learning visual representations from text supervision have shown tremendous success on transferring to downstream tasks~\cite{radford2021learning}. The learned model can not only be transferred to ImageNet classification in a zero-shot manner and achieve state-of-the-art performance, but can also perform recognition on object categories beyond ImageNet.  
Inspired by this line of research, we ask the question: Can we also learn a semantic segmentation model purely with text supervision, and without any per-pixel annotations, capable of generalizing to different sets of objects categories, or vocabularies, in a zero-shot manner? 

To accomplish this, we propose to bring back the grouping mechanism into deep networks, which allows semantic segments to emerge automatically with only text supervision. An overview of our approach is illustrated in Fig.~\ref{fig:teaser}. By training on large-scale paired image-text data with contrastive losses, we enable the model to be zero-shot transferred to several semantic segmentation vocabularies, without requiring any further annotation or fine-tuning. Our key idea is to leverage the Vision Transformer (ViT)~\cite{dosovitskiy2020image} and incorporate a new visual grouping module into it. 

We call our model \emph{GroupViT} (Grouping Vision Transformer). Compared to convolutional neural networks (ConvNets), which operate on regular grids, the global self-attention mechanism of Transformers naturally provides the flexibility to combine visual tokens into non-grid-like segments. Thus, instead of organizing visual tokens into grids, as recent ViT-based applications~\cite{liu2021swin, chu2021twins, el2021xcit, wu2021cvt} do, we propose to perform hierarchical grouping of visual tokens into irregular-shaped segments. Specifically, our GroupViT model is organized in different stages through a hierarchy of Transformer layers, where each stage contains multiple Transformers to perform information propagation among the group segments, and a grouping module that merges smaller segments into larger ones. With different input images, our model dynamically forms different visual segments, each intuitively representing a semantic concept. 

We train GroupViT with text supervision only. To perform learning, we merge visual segment outputs in the final stage of GroupViT using average pooling. We then compare this image-level embedding to those derived from textual sentences via contrastive learning. We construct positive training pairs by using corresponding image and text pairs, and negative ones by using text from other images. We extract the text embedding with a Transformer model, trained jointly along with GroupViT from scratch. Interestingly, even though we only provide textual training supervision at the image level, we find that semantically meaningful segments automatically emerge using our grouping architecture.

During inference, for the task of semantic segmentation, given an input image, we extract its visual groups using GroupViT (Fig.~\ref{fig:teaser}). Each final group's output represents a segment of the image. Given a vocabulary of label names for segmentation, we use the text Transformer to extract each label's textual embedding. To perform semantic segmentation, we then assign the category labels to image segments according to their mutual similarity in the embedding space. In our experiments, we show that GrouViT trained on the Conceptual Caption~\cite{sharma2018conceptual, changpinyo2021conceptual} and Yahoo Flickr Creative Commons~\cite{thomee2016yfcc100m} datasets with text supervision alone, can transfer to semantic segmentation tasks on the PASCAL VOC~\cite{everingham2010pascal} and PASCAL Context~\cite{mottaghi2014role} datasets in a zero-shot manner. Without any fine-tuning, we achieve a mean intersection over union (mIoU) of 52.3\%  on PASCAL VOC 2012 and an mIoU of 22.4\% on PASCAL Context, performing competitively to state-of-the-art transfer-learning methods requiring greater levels of supervision.
To our knowledge, our work is the first to perform semantic segmentation on different label vocabularies in a zero-shot manner with text supervision alone, without requiring any pixel-wise labels.

Our contributions are the following:
\begin{itemize}[noitemsep,nosep]
    \item Moving beyond regular-shaped image grids in deep networks, we introduce a novel GroupViT architecture to perform hierarchical bottom-up grouping of visual concepts into irregular-shaped groups.
    \item Without any pixel-level labels and training and with only image-level text supervision using contrastive losses, GroupViT successfully learns to group image regions together and transfers to several semantic segmentation vocabularies in a zero-shot manner.
    \item To our knowledge, ours is the first work to explore zero-shot transfer from text supervision alone to several semantic segmentation tasks without using \emph{any} pixel-wise labels and establishes a strong baseline for this new task.
\end{itemize}

\section{Related Work}
\vspace{-0.25em}
\noindent\textbf{Vision Transformer.} Inspired by the success of Transformers in NLP~\cite{vaswani2017attention,devlin2018bert}, the Vision Transformer (ViT)~\cite{dosovitskiy2020image} was recently proposed and has been successfully applied to multiple computer vision tasks, including image classification~\cite{touvron2021training,liu2021swin,yuan2021tokens, touvron2021going}, object detection~\cite{liu2021swin, zhang2021multi, wang2021pyramid}, semantic segmentation~\cite{zheng2021rethinking, liu2021swin, xie2021segformer} and action recognition~\cite{bertasius2021space,fan2021multiscale, liu2021video, ryoo2021tokenlearner, arnab2021vivit}. However, much like ConvNets, most variants of ViT still operate on regular image grids. For example, Liu et al.~\cite{liu2021swin} divide the image into regular shaped windows and apply a Transformer block to each one. The convolutional operations are also inserted back into the Transformer block in~\cite{chu2021twins, el2021xcit, wu2021cvt}. While these variants of ViT achieve remarkable performance, they don't fully leverage the flexibility of the global self-attention mechanism in Transformers. 
That is, self-attention, by design, can be applied to any arbitrary image segments and is not limited to rectangular-shaped and scan-ordered ones only. Our GroupViT model, on the other hand, leverages this property of Transformers to learn to group visual information into several arbitrary-shaped segments. With a hierarchical design, it further merges smaller segments into larger ones and yields different semantic groups for each image.

\noindent\textbf{Representation Learning with Text Supervision.} With large-scale image-text paired data available on the Internet, representation learning with text supervision~\cite{joulin2016learning,lu2019vilbert,li2019visualbert,chen2019uniter,li2020oscar,desai2021virtex,zhang2020contrastive,radford2021learning,jia2021scaling} has been shown to be successful in transferring to various down stream tasks such as visual question answering~\cite{antol2015vqa,zhou2020unified} and visual reasoning~\cite{zellers2019recognition}. For example, Desai et al.~\cite{desai2021virtex} pre-train ConvNets with the image captioning task, and transfer the representation by fine-tuning with downstream task annotations, e.g., object detection labels. Recently, Radford et al.~\cite{radford2021learning} propose to perform contrastive learning between image and text. They show that the learned model can be directly transferred to ImageNet classification~\cite{deng2009imagenet} in a zero-shot manner without any fine-tuning. Going beyond image classification, our GroupViT model further explores zero-shot transfer to semantic segmentation tasks with only text supervision, which has not been shown in previous work to the best of our knowledge.

\noindent\textbf{Visual Grounding.} Visual grounding aims to learn image region-text correspondence. One line of research explores a fully supervised approach to detecting text-related bounding boxes within an image~\cite{plummer2018conditional, chen2019uniter, lu2019vilbert, gan2020large, kamath2021mdetr} using datasets such as Flickr30k Entities~\cite{plummer2015flickr30k} and Visual Genome ~\cite{krishna2017visual}. To scale up learning, weakly-supervised visual grounding has been introduced where the bounding box and text correspondence is not available during training~\cite{wang2021improving, yeh2018unsupervised, liu2021relation, gupta2020contrastive, liu2019knowledge, chen2018knowledge}. However, to localize object bounding boxes these approaches still rely on pre-trained object detectors~\cite{wang2021improving, yeh2018unsupervised}, which, in turn, utilize box annotations from other datasets. 
While related, we emphasize there are two main differences between our problem setting and that of visual grounding: (i) We train our model on millions of noisy image-text pairs from the web, while visual grounding 
requires human curated and annotated data at a relatively smaller scale; (ii) Our GroupViT provides a bottom-up mechanism for progressive visual grouping where object segments automatically emerge with text supervision, while visual grounding 
needs bounding box annotations borrowed from other datasets. 

\noindent\textbf{Semantic Segmentation with Less Supervision.} Multiple research directions have been proposed to learn to segment with less supervision than dense per-pixel labels. For example, few-shot learning~\cite{dong2018few, yang2021mining, lu2021simpler, tian2020prior, wang2020few, nguyen2019feature, liu2020part} and active learning~\cite{shin2021all, casanova2020reinforced, xie2020deal, sinha2019variational, sener2017active} are proposed to perform segmentation with as few pixel-wise labels as possible. Going further, zero-shot approaches~\cite{bucher2019zero,li2020consistent} are proposed to learn segmentation models for unseen categories without using pixel-wise labels for them. However, it still requires learning with segmentation labels on seen categories as the initial step. Another line of related research is of weakly-supervised semantic segmentation~\cite{ahn2018learning, li2021pseudo, jo2021puzzle, sun2020mining, wang2020self, fan2020learning, lee2019ficklenet, chang2020weakly, shimoda2019self}, which aims to learn semantic segmentation with only image-level object category supervision. While it largely reduces supervision, it still requires manual labeling using a finite vocabulary on a carefully-curated image dataset. Different from all previous work, our approach completely gets rid of human annotations and GroupViT is trained with large-scale noisy text supervision. Instead of a fixed vocabulary, we show that GroupViT can be generalized to any set of categories in a zero-shot manner for semantic segmentation.

The concurrently developed unpublished text-supervised semantic segmentation methods~\cite{zhou2021denseclip, xu2021simple, zabari2021semantic, ghiasi2021open} also show promising results. 
One major difference between these methods and GroupViT is that, they exploit vision-language model \cite{radford2021learning, jia2021scaling} pre-trained on well-prepared large-scale private dataset with 400M-1.8B image-text pairs, while our GroupViT is trained from scratch with much noisier public datasets (30M images in total) to learn grouping and segmentation and yet achieves competitive performance. 
Among these works, OpenSeg~\cite{ghiasi2021open} also learns with class agnostic mask annotations to generate mask proposals, while our method does not require any mask annotations.


\begin{figure*}[t]
    \centering
    \vspace{-.5em}
    \includegraphics[width=.99\linewidth]{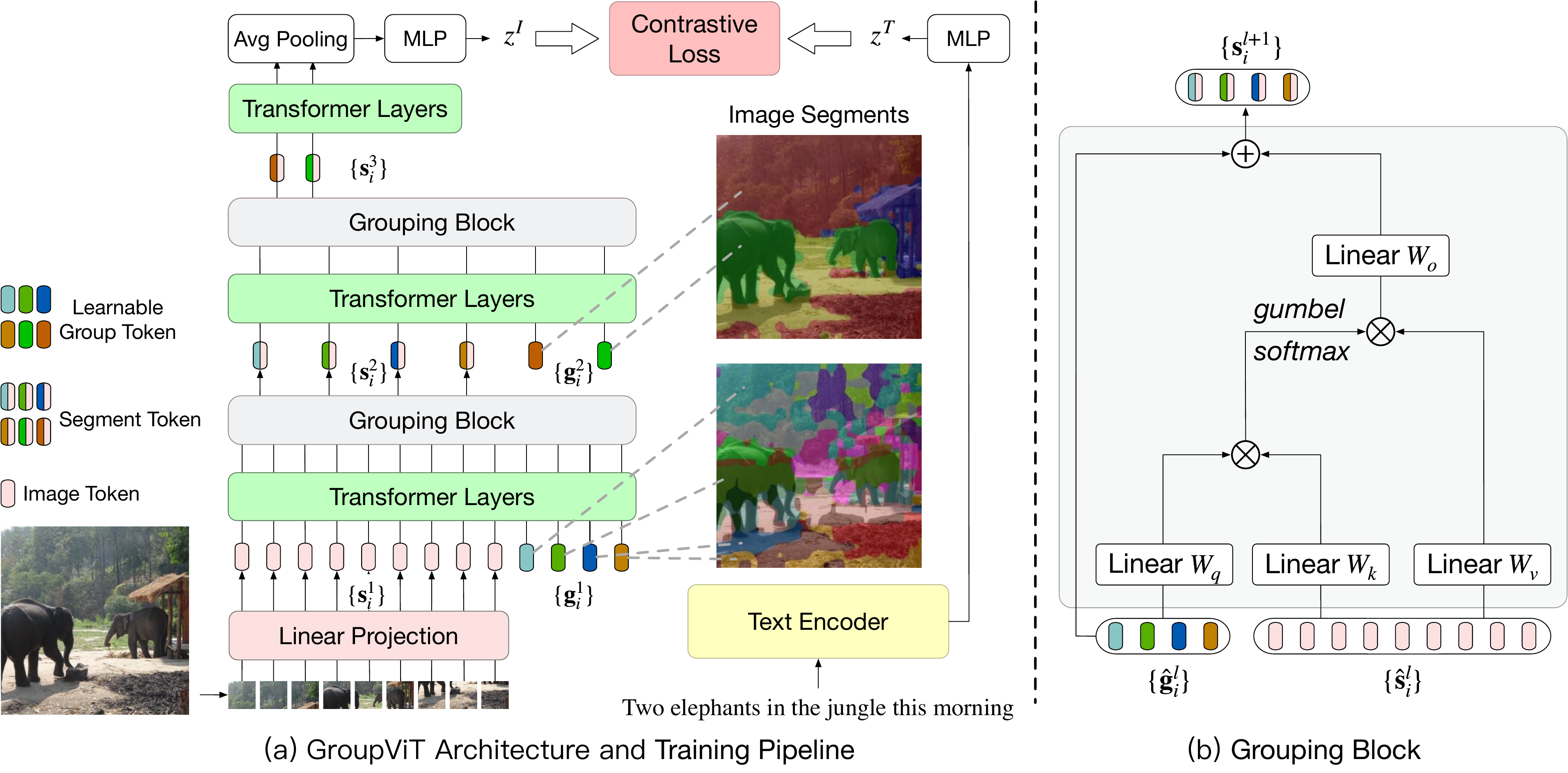}
    \vspace{-.5em}
    \caption{
        (a) \textbf{The Architecture and Training Pipeline of GroupViT.} GroupViT contains a hierarchy of Transformer layers grouped into stages, each operating on progressively larger visual segments.
        The images on the right show visual segments that emerge in the different grouping stages.
        The lower stage groups pixels into object parts, e.g., noses and legs of elephants; and the higher stage further merges them into entire objects, e.g., the whole elephant and the background forest.
        (b) \textbf{The Architecture of Grouping Block}. 
        Each grouping stage ends with a grouping block that computes the similarity between the learned group tokens and segment (image) tokens. 
        The assignment is computed via gumbel softmax over group tokens and converted into a one-hot hard assignment.
        The segment tokens assigned to the same group are merged together and represent new segment tokens that are input to the next grouping stage.
    }
    \label{fig:pipeline}
    \vspace{-1.5em}
\end{figure*}

\section{Method}
\vspace{-0.5em}
We propose the GroupViT architecture for zero-shot transfer to semantic segmentation with text supervision only. GroupViT introduces a new hierarchical grouping Transformer architecture that exploits the global self-attention mechanism of Transformers to partition input images into progressively larger arbitrary-shaped groups. 
We first describe GroupViT's architecture in detail in Sec.~\ref{sec:grouping-vision-transformer}. To train it, we employ carefully-designed contrastive losses between image-text pairs, 
which we describe in Sec.~\ref{sec:multilabel}. Lastly, we transfer the trained GroupViT model, without further fine-tuning, to the task of zero-shot semantic segmentation as described in Sec.~\ref{sec:zero-shot-transfer}. 
\subsection{Grouping Vision Transformer}
\vspace{-0.25em}
\label{sec:grouping-vision-transformer}
We introduce the GroupViT image encoder (Fig.~\ref{fig:pipeline}), which performs hierarchical progressive grouping of visual concepts via a Transformer-based architecture. In GroupViT, we separate Transformer layers into multiple grouping stages. In each stage, we learn a number of group tokens (as learnable parameters) via self-attention that aggregate information globally from all image tokens (segments). We then use the learned group tokens to merge similar image tokens together via a \textit{Grouping Block}. Through a hierarchy of grouping stages, we group smaller image segments into larger ones. We describe each component next.

\paragraph{Architecture}
Following the design of ViT~\cite{dosovitskiy2020image}, we first split an input image into $N$ non-overlapping patches and linearly project each into a latent space. We treat each projected patch as an input image token and denote the set of all of them as $\{{\bf{p}}_i\}_{i=1}^N$. In each grouping stage, besides the image tokens, we concatenate a set of learnable group tokens and input them into the Transformer for that stage.

\paragraph{Multi-stage Grouping}
As Fig.~\ref{fig:pipeline}(a) shows, instead of forwarding all the $N$ input image tokens through all the layers of the Transformer, we separate its layers into a hierarchy of grouping stages. 
Each stage incorporates a \textit{Grouping Block} at its end to merge the smaller groups into larger ones.

Formally, suppose there are $L$ grouping stages, each indexed by $l$ and with a set of learnable group tokens $\{{\bf{g}}_i\}_{i=1}^{M_l}$.
For simplicity, we treat the image patches $\{{\bf{p}}_i\}_{i=1}^N$ input to the first grouping stage as the set of starting segments $\{{\bf{s}}^1_i\}_{i=1}^{M_0}$ , where $N = M_0$.
We simplify $\{{\bf{s}}^l_i\}_{i=1}^{M_{l-1}}$ to $\{{\bf{s}}^l_i\}$ and similarly $\{{\bf{g}}^l_i\}_{i=1}^{M_{l}}$ to $\{{\bf{g}}^l_i\}$.
Starting with $l{=}1$, for each grouping stage, we first concatenate $\{{\bf{s}}^l_i\}$ and $\{{\bf{g}}_i^{l}\}$ together and then input them into a number of Transformer layers, each of which performs information propagation between them via
\begin{equation*}
\{\hat{\bf{g}}_i^{l}\}, \{\hat{\bf{s}}^l_i\} = {\text{Transformer}}([\{{\bf{g}}_i^{l}\};\{{\bf{s}}^l_i\}]),
\end{equation*}
where $[~;~]$ denotes the concatenation operator. 
Then we group the updated $M_{l-1}$ image segment tokens $\{\hat{\bf{s}}^l_i\}$ into $M_{l}$ new segment tokens $\{{\bf{s}}^{l+1}_i\}_{i=1}$ via a Grouping Block as
\begin{equation*}
\{{\bf{s}}^{l+1}_i\} = \text{GroupingBlock}(\{\hat{\bf{g}}_i^{l}\}, \{\hat{\bf{s}}^l_i\}).
\end{equation*}
In each grouping stage $M_{l} <  M_{l-1}$, i.e., there are progressively fewer group tokens, resulting in progressively larger and fewer image segments.
After the final grouping stage, $L$, we apply Transformer layers on all segment tokens and finally average their outputs to obtain the final global image representation $z^I$ as
\vspace{-.5em}
\begin{align}
    &\{\hat{\bf{s}}^{L+1}_i\} = \text{Transformer}(\{{\bf{s}}^{L+1}_i\}),\\
    &z^I = \text{MLP}(\text{AvgPool}(\{\hat{\bf{s}}^{L+1}_i\})).
\vspace{-.5em}
\label{eqn:final_image}
\end{align}
As shown in Fig.~\ref{fig:pipeline}(a), GroupViT re-organizes visual information into arbitrary image segments after the first stage itself and thus is not confined to a regular-grid structure.

\paragraph{Grouping Block} 
\label{sec:grouping-block}
As shown in Fig.~\ref{fig:pipeline}(b), the Grouping Block at the end of each grouping stage takes the learned group tokens and image segment tokens as inputs. It merges all the segment tokens that are assigned to the same group token into a single new image segment, based on similarity in the embedding space.

Formally, we compute the similarity matrix $\mathbf{A}^l$ between the group tokens $\{{\hat{\bf{g}}}_i^{l}\}$ and segment tokens $\{\hat{\bf{s}}_i^{l}\}$ via a \texttt{Gumbel-Softmax}~\cite{jang2016categorical, maddison2016concrete} operation computed over the group tokens as
\vspace{-1em}
\begin{equation}
    \mathbf{A}_{i, j}^l = \frac{\exp(W_q\hat{\bf{g}}_i^{l} {\cdot} W_k\hat{\bf{s}}_j^l+\gamma_i)}{\sum_{k=1}^{M_{l}} \exp(W_q\hat{\bf{g}}_k^l {\cdot} W_k\hat{\bf{s}}_j^l+\gamma_k)},
\vspace{-.5em}
\end{equation}
where $W_q$ and $W_k$ are the weights of the learned linear projections for the group and segment tokens, respectively, and $\{\gamma_i\}$ are i.i.d random samples drawn from the \texttt{Gumbel(0, 1)} distribution.
We compute the group to assign a segment token to by taking the one-hot operation of it \texttt{argmax} over all the groups.
Since the one-hot assignment operation via \texttt{argmax} is not differentiable, we instead use the straight through trick in~\cite{oord2017neural} to compute the assignment matrix as
\vspace{-.5em}
\begin{equation}
\hat{\mathbf{A}}^l = \texttt{one-hot}({\mathbf{A}}^l_{\texttt{argmax}}) + \mathbf{A}^l - \texttt{sg}(\mathbf{A}^l),
\vspace{-.5em}
\end{equation}
where $\texttt{sg}$ is the stop gradient operator. 
With the straight through trick, $\hat{\mathbf{A}}^l$ has the one-hot value of assignment to a single group, but its gradient is equal to the gradient of $\mathbf{A}^l$, which makes the Grouping Block differentiable and end-to-end trainable.
We call this one-hot assignment strategy as \textit{hard assignment}.
After assigning the segment tokens to the different learned groups, we merge the embedding of all the tokens belonging to the same group to form a new segment token ${\bf{s}}_i^{l+1}$.
For each group, the output of the Grouping Block is a weighted sum of the segment tokens assigned to that group and computed as
\vspace{-1em}
\begin{equation}
\label{eqn:assign}
    {\bf{s}}_i^{l+1} = \hat{\bf{g}}_i^{l} + W_o\frac{\sum_{j=1}^{M_{l-1}}\hat{\mathbf{A}}_{i, j}^{l}W_v\hat{\bf{s}}_j^l}{\sum_{j=1}^{M_{l-1}} \hat{\mathbf{A}}_{i, j}^{l}},
\end{equation}
where $W_v$ and $W_{o}$ are the learned weights to project the merged features. An alternative to hard assignment is soft assignment, which uses $\mathbf{A}^l$ instead of $\mathbf{\hat{A}}^l$ for computing Eqn.~\ref{eqn:assign}.
Empirically, we found that hard assignment results in more effective grouping versus soft assignment (Table~\ref{tab:hard-multi}).

The Grouping Block works similarly to a single iteration of the previously proposed Slot Attention mechanism~\cite{locatello2020object}. While Slot Attention learns instance-level grouping from self-supervision, our Grouping Block groups similar semantic regions with weak text supervision.  
For example, in the second row of Fig.~\ref{fig:vis_small}, the two horses are grouped together. 

\subsection{Learning from Image-Text Pairs}
To train GroupViT to perform hierarchical grouping, we employ carefully-designed contrastive losses between image-text pairs. We describe these next.

\paragraph{Image-Text Contrastive Loss}
To learn visual representations via text supervision, following~\cite{radford2021learning, jia2021scaling}, we train a dual-encoder architecture via an image-text contrastive loss. In our case, GroupViT acts as the image encoder and a Transformer~\cite{vaswani2017attention} as the text encoder.
The final image embedding from GroupViT (Eqn.~\ref{eqn:final_image}) is the average embedding of all its output segment tokens.
The text embedding is the embedding of the last output token (end-of-sentence token) from the text Transformer.
We forward the input image and text in a pair through their respective encoders, project them into a common embedding space and compute a similarity measure between them.
We consider all matched image-text pairs as positive pairs, and all other unmatched ones as negative ones.
Our training objective is to pull the representations of the positive pairs closer to each other, while pushing those of the unmatched ones far away from each other.

Formally, assume a batch of $B$ image-text pairs $\{(x_i^I, x_i^T)\}_{i=1}^B$, where $x_i^I$ and $x_i^T$ are the image and text inputs, respectively, of the $i$-th pair. We encode each of them, via their respective encoders, into embedding vectors $z_i^I$ and $z_i^T$ and $l_2$-normalize each.
We then measure their similarity by computing their dot product. The total image-text contrastive loss is defined as
\vspace{-1em}
\begin{equation}
\mathcal{L}_{I \leftrightarrow T} = \mathcal{L}_{I \rightarrow T}+\mathcal{L}_{T \rightarrow I},
\label{eqn:i2t}
\vspace{-.5em}
\end{equation}
which
is composed of an image-to-text contrastive loss defined as
\vspace{-1em}
\begin{equation*}
\mathcal{L}_{I \rightarrow T} = -\frac{1}{B}\sum_{i=1}^B\log \frac{\exp(z_i^I {\cdot} z_i^T/\tau)}{\sum_{j=1}^B \exp(z_i^I {\cdot} z_j^T/\tau)},
\vspace{-.5em}
\end{equation*}
and a text-to-image contrastive loss defined as
\vspace{-1em}
\begin{equation*}
\mathcal{L}_{T \rightarrow I} = -\frac{1}{B}\sum_{i=1}^B\log \frac{\exp(z_i^T {\cdot} z_i^I/\tau)}{\sum_{j=1}^B \exp(z_i^T {\cdot} z_j^I/\tau)},
\vspace{-.5em}
\end{equation*}
where $\tau$ is a learnable temperature parameter to scale the logits.

\begin{figure}[t]
    \centering
    \includegraphics[width=.99\linewidth]{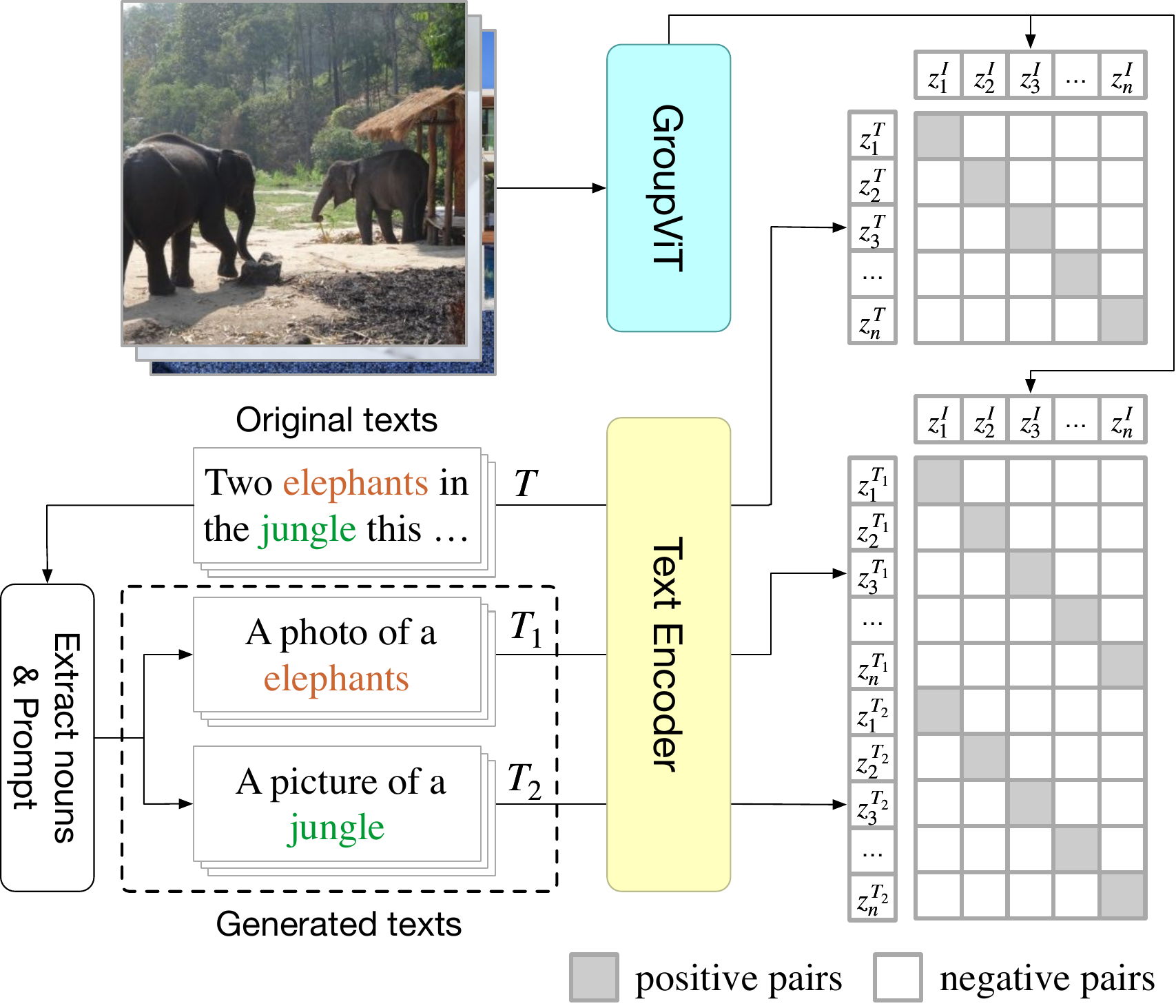}
    \vspace{-.5em}
    \caption{
        \textbf{Multi-label Image-Text Contrastive Loss}. 
        Given an input image-text pair, we generate new text from the original text by extracting its nouns and by prompting them with several sentence templates. 
        For constrastive learning, we treat only matched image and text pairs as positive ones.
        We train GroupViT and the text encoder to maximize the feature similarity between positive image-text pairs and to minimize it between negative pairs. 
    }
    \label{fig:multilabel}
    \vspace{-1.75em}
\end{figure}

\paragraph{Multi-Label Image-Text Contrastive Loss}
\label{sec:multilabel}
To enable effective visual grouping, 
besides the image-text loss in Eqn.~\ref{eqn:i2t}, we propose a multi-label contrastive loss with text prompting.
As illustrated in Fig.~\ref{fig:multilabel},
we use the ``prompting engineering" mechanism proposed in ~\cite{radford2021learning} to generate additional text labels for each image besides its originally provided sentence label. 
Specifically, we randomly select $K$ noun words from a sentence $x_i^{T}$, and prompt each of them with a set of handcrafted sentence templates, e.g., ``\texttt{A photo of a \{noun\}}".
The motivation to select nouns is that objects in images are more likely to be described by them. 
Besides training with the original image-text pairs $\{(x_i^I, x_i^T)\}_{i=1}^B$, 
we employ additional contrastive losses between the new sets of image-``prompted text" pairs $\{(x_i^I, x_i^{T_1})\}_{i=1}^B, \{(x_i^I, x_i^{T_2})\}_{i=1}^B, \dots, \{(x_i^I, x_i^{T_K})\}_{i=1}^B$, 
where $\{x_i^{T_k}\}_{k=1}^K$ are all prompted sentences generated from the nouns sampled from $x_i^{T}$. 
As shown in Fig.~\ref{fig:multilabel}, compared to the standard contrastive loss (Eqn.~\ref{eqn:i2t}), which results in only one positive pair among the batch $B$, in our case, each image $x_i^{I}$ has $K$ positive text pairs and $B(K-1)$ negative ones.

Similarly to the standard image-text contrastive loss (Eqn.~\ref{eqn:i2t}), our multi-label contrastive loss is defined as
\vspace{-.5em}
\begin{equation}
\mathcal{L}_{I \leftrightarrow \{T_k\}_{k=1}^K} = \mathcal{L}_{I \rightarrow \{T_k\}_{k=1}^K}+\mathcal{L}_{\{T_k\}_{k=1}^K \rightarrow I},
\end{equation}
which is a sum of two two-way contrastive losses
\vspace{-.5em}
\begin{equation*}
\mathcal{L}_{I \rightarrow \{T_k\}_{k=1}^K} = -\frac{1}{B}\sum_{i=1}^B\log \frac{\sum_{k=1}^K\exp(z_i^I {\cdot} z_i^{T_k}/\tau)}{\sum_{k=1}^K \sum_{j=1}^B \exp(z_i^I {\cdot} z_j^{T_k}/\tau)}
\vspace{-.5em}
\end{equation*}
and
\vspace{-1em}
\begin{equation*}
\mathcal{L}_{\{T_k\}_{k=1}^K \rightarrow I} = -\frac{1}{KB}\sum_{k=1}^K\sum_{i=1}^B\log \frac{\exp(z_i^{T_k} {\cdot} z_i^I/\tau)}{\sum_{j=1}^B \exp(z_i^{T_k} {\cdot} z_j^I/\tau)}.
\vspace{-.5em}
\end{equation*}
Finally, the total image-text contrastive loss for training GroupVIT is defined as
\vspace{-.5em}
\begin{equation}
\mathcal{L} = \mathcal{L}_{I \leftrightarrow T} + \mathcal{L}_{I \leftrightarrow \{T_k\}_{k=1}^K}.
\label{eqn:multilabel}
\end{equation}

\subsection{Zero-Shot Transfer to Semantic Segmentation}
\label{sec:zero-shot-transfer}
\begin{figure}[t]
    \centering
    \includegraphics[width=.99\linewidth]{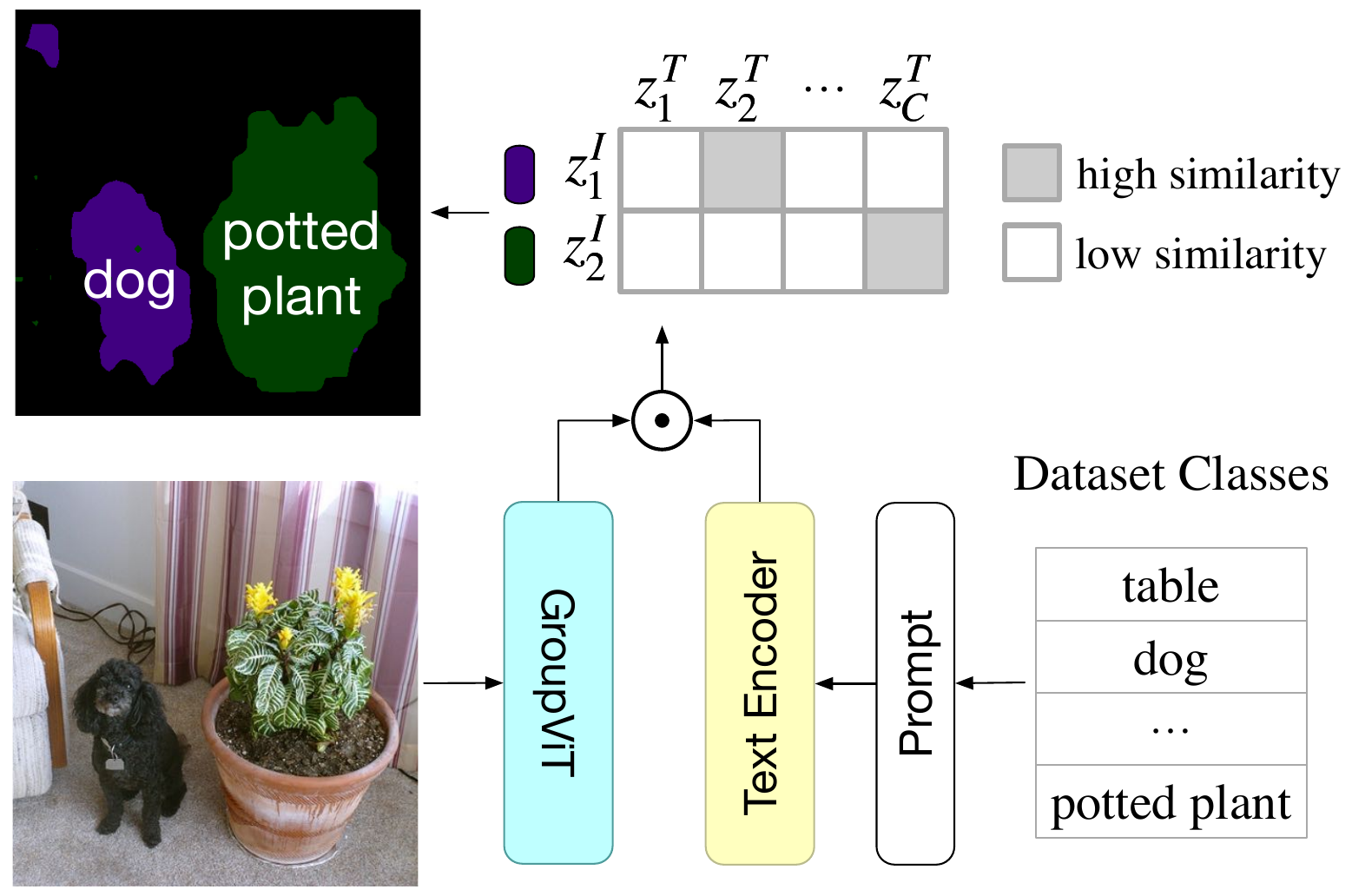}
    \vspace{-.5em}
    \caption{
    \textbf{Zero-Shot Transfer of GroupViT to Semantic Segmentation}.
    Each output segment's embedding from GroupViT corresponds to a region of the image. We assign each output segment to the object class with the highest image-text similarity in the embedding space.
    }
    \label{fig:inference}
    \vspace{-1.5em}
\end{figure}

Since GroupViT automatically groups images into semantically-similar segments, its output can be easily zero-shot transferred to semantic segmentation without any further fine-tuning.
This zero-shot transfer pipeline is illustrated in Fig.~\ref{fig:inference}.
To infer the segments of an image belonging to a finite vocabulary of object classes, we forward a test image through GroupVIT without applying \texttt{AvgPool} to its final $L$ output segments, and obtain the embedding of each of them as $\{z^I_i\}^{M_{L}}_{i=1}$. 
Each segment token corresponds to an arbitrarily-shaped region of the input image. We then compute the similarity between the embedding of each segment token and the text embedding of all the semantic classes present in the dataset. 
We assign each image segment to the semantic class with the highest image-text embedding similarity. 

Specifically, let $\hat{\mathbf{A}}^l$ be the assignment matrix of the $l$-th grouping stage described in Sec.~\ref{sec:grouping-block}, which indicates the mapping between the input and output segments of $l$-th stage.
Multiplying all the stage-level assignment matrices $\prod_{l=L}^1 \hat{\mathbf{A}}^l$ yields the final assignment between the input patches $\{{\bf{p}}_i\}_{i=1}^N$ and the final-stage output tokens $\{z^I_i\}^{M_{L}}_{i=1}$.
We use the same ``prompting engineering" as described in Sec.~\ref{sec:multilabel} to transform all the semantic segmentation label names into sentences.
The embedding of label names in the dataset is $\{z^{T}_i\}_{i=1}^C$, where $C$ is the number of classes. 
As shown in Fig.~\ref{fig:inference}, to classify an image segment $z^{I}_i$ to one of $C$ classes, we compute the dot product between $l_2$-normalized class name embedding vectors $\{z^{T}_i\}_{i=1}^C$ and $z^{I}_i$, and assign it to the class with the highest similarity.

\section{Experiments}
\subsection{Implementation Details}
\paragraph{Architecture}
The architecture of GroupViT is based on ViT-S~\cite{dosovitskiy2020image, touvron2021training} with 12 Transformer layers, each with a hidden dimension of 384. We use input images of size 224 $\times$ 224 and a patch size of 16 $\times$ 16. We add a learnable positional embedding to each patch after linearly projecting it. We experiment with 1-stage and 2-stage architectures for GroupViT. 
Both architectures output 8 tokens after the final grouping stage. 
In 1-stage GroupViT, we learn 64 group tokens and insert the grouping block after the sixth Transformer layer.
Before the grouping block, we project the 64 group tokens into 8 tokens using an MLP-Mixer layer~\cite{tolstikhin2021mlp} and output 8 segment tokens. 
In 2-stage GroupViT, there are 64 and 8 group tokens in the first and second grouping stages, respectively.
We insert grouping blocks after the sixth and ninth Transformer layers.
Our text-encoder is the same as \cite{radford2021learning}.
We use a 2-layer MLP to project the visual and text embedding vectors into the same latent space.
\paragraph{Training}
We use the CC12M~\cite{changpinyo2021conceptual} and the filtered YFCC~\cite{thomee2016yfcc100m} datasets for training, containing 12M and 14M image-text pairs, respectively.
Our batch size is 4096 with a learning rate initialized to 0.0016 and decayed via the cosine schedule~\cite{loshchilov2016sgdr}.
We use the Adam~\cite{kingma2014adam} optimizer with a weight decay of 0.05. 
We train GroupVIT for 30 epochs with the 5 initial epochs containing linear warm-up. 
For the multi-label contrastive loss, we set $K=3$. We use the same text templates as in~\cite{radford2021learning} for generating text prompts. 
\paragraph{Zero-shot Transfer to Semantic Segmentation}
We evaluate GroupViT for the task of zero-shot transfer to semantic segmentation on the validation splits of the PASCAL VOC 2012~\cite{everingham2010pascal} and PASCAL Context~\cite{mottaghi2014role} datasets. 
They each contain 20 and 59 foreground classes, respectively, with an additional background class.
During inference, GroupViT predicts only the foreground classes by thresholding the softmax-normalized-similarity between the embedding of the output image segments and the text segmentation labels, where we set the threshold to 0.9 and 0.5 for PASCAL VOC 2012 and PASCAL Context, respectively.
We resize each input image to have a shorter side length of 448.

\subsection{Ablation Study}
\label{sec:ablation}
To discern the contribution of each component of GroupViT, we conduct an ablation study. 
For all experiments, we train a 1-stage GoupViT with the CC12M dataset, unless otherwise specified.
We report mIoU (mean intersection over union) of the predicted and ground truth segmentation masks on the PASCAL VOC 2012 validation set.

\paragraph{Hard vs. Soft Assignment}
In each Grouping Block, we assign image segment tokens to group tokens via hard or soft assignment (Sec.~\ref{sec:grouping-block}). For soft assignment, we use the original ${\bf{A}}^l$ matrix instead of $\hat{\bf{A}}^l$ used for hard assignment to compute Eqn.~\ref{eqn:assign}. 
The impact of this is shown in the first column of Table~\ref{tab:hard-multi}. 
We find that hard assignment improves over soft assignment by a large margin, $>$10\% mIoU. 
We conjecture that with soft assignment, the features of new segment tokens $\{{\bf{s}}_i^{l+1}\}$ are likely to be more correlated with each other due to absence of zero values in ${\bf{A}}^l$.
Hence, each group may contain information from the same image patches increasing ambiguity while assigning text labels to image segments. 
With hard assignment, however, the affinity matrix $\hat{\bf{A}}^l$ assigns image segments to groups in a mutually exclusive manner, making groups more differentiated and their assignment to text labels less ambiguous.

\begin{table}[]
  \tablestyle{6pt}{1.1}
  \begin{tabular}{c|cc|c}
  arch        
  & {\tablestyle{0pt}{.9} \begin{tabular}{c} {hard} \\ {assignment} \end{tabular}}  
  & {\tablestyle{0pt}{.9} \begin{tabular}{c} {multi-} \\ {label loss} \end{tabular}}  
  & {\tablestyle{0pt}{.9} \begin{tabular}{c} {mask} \\ {mIoU} \end{tabular}}  \\
  \shline
  GroupViT &             &             & 12.0 \\
  GroupViT & \cmark      &             & 36.7 \\
  GroupViT &             & \cmark      & 25.1 \\
  GroupViT & \cmark      & \cmark      & \textbf{39.3}
  \end{tabular}
  \vspace{-.5em}
  \caption{
    \textbf{Ablation results of hard vs. soft assignment and multi-label contrastive loss}. 
  }
  \label{tab:hard-multi}
  \vspace{-1.5em}
\end{table}

\begin{table}[]
  \tablestyle{6pt}{1.1}
  \begin{tabular}{c|cc|cc}
  arch        
  & {\tablestyle{0pt}{.9} \begin{tabular}{c} {\# group} \\ {tokens} \end{tabular}}  
  & {\tablestyle{0pt}{.9} \begin{tabular}{c} {\# output} \\ {tokens} \end{tabular}}  
  & {\tablestyle{0pt}{.9} \begin{tabular}{c} {mask} \\ {mIoU} \end{tabular}}  \\
  \shline
  GroupViT & 16          & 4             & 28.6 \\
  GroupViT & 16          & 8             & 37.1 \\
  GroupViT & 32          & 8             & 38.3 \\
  GroupViT & 64          & 8             & \textbf{39.3} \\
  GroupViT & 64          & 16            & 38.0 \\
  \end{tabular}
  \vspace{-.5em}
  \caption{
    \textbf{Ablation results of different numbers of group and output tokens}. 
  }
  \label{tab:group-num}
  \vspace{-1.5em}
\end{table}

\begin{table}[]
  \tablestyle{5pt}{1.1}
  \begin{tabular}{c|cc|cc}
  arch        
  & dataset              
  & {\tablestyle{0pt}{.9} \begin{tabular}{c} {\#} \\ {stages} \end{tabular}}  
  & {\tablestyle{0pt}{.9} \begin{tabular}{c} {mask} \\ {mIoU} \end{tabular}}  
  & {\tablestyle{0pt}{.9} \begin{tabular}{c} {boundary} \\ {mIoU} \end{tabular}}  \\
  \shline
  GroupViT & CC12M      & 1      & 39.3 & 31.6     \\
  GroupViT & CC12M      & 2      & 41.1 & 33.5     \\
  \hline
  GroupViT & CC12M+YFCC & 1      & 37.2 & 32.3     \\
  GroupViT & CC12M+YFCC & 2      & \textbf{52.3} & \textbf{40.3}
  \end{tabular}
  \vspace{-.5em}
  \caption{
    \textbf{Ablation results of single-stage and multi-stage grouping}. 
  }
  \label{tab:multi-stage}
  \vspace{-1em}
\end{table}

\begin{figure}[]
    \centering
    \includegraphics[width=.9\linewidth]{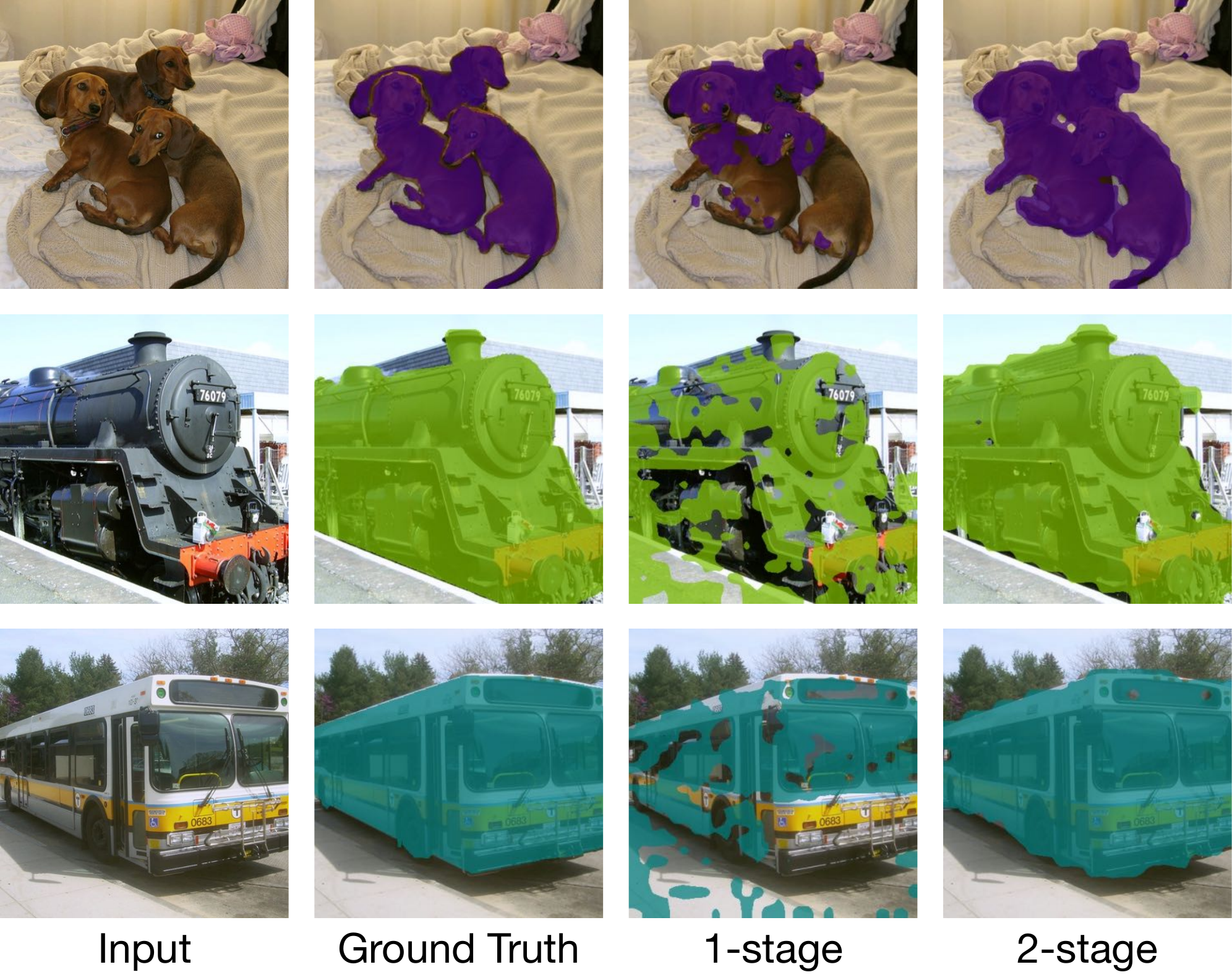}
    \vspace{-0.5em}
    \caption{
    \textbf{Visual results of 1-stage and 2-stage GroupViT}. 
    The segmentation maps generated by 2-stage GroupViT are smoother and more accurate than those of 1-stage GroupViT. 
    }
    \label{fig:vis_stage}
    \vspace{-1.em}
\end{figure}

\paragraph{Multi Label Contrastive Loss}
We investigate the effect of adding the multi-label contrastive loss in the second column of Table~\ref{tab:hard-multi}.
Adding the multi-label contrastive loss to the standard one (Eqn.~\ref{eqn:multilabel}) improves performance both with hard and soft assignment, by 13.1\% and 2.6\%, respectively.
With the multi-label contrastive loss, the input text during training and inference is in a similar prompted format. 
We conjecture that this consistency helps GroupViT better classify the learned image segments into label categories.

\paragraph{Group Tokens}
In Table~\ref{tab:group-num}, we compare different group and output tokens. We observe that increasing group tokens consistently improves performance. Conceptually, each group token represents a distinct semantic concept. 
So more group tokens presumably help GroupViT learn to group more semantic concepts.
Note that although the number of group tokens is much less than the number of classes in the real world, each group token is a feature vector in a 384-D embedding space, which can represent many more concepts than just 1.
We also experiment with different output tokens and find 8 to be optimal, similar to findings in \cite{ryoo2021tokenlearner}. 

\begin{table}[]
  \tablestyle{8pt}{1.1}
  \begin{tabular}{c|c|ccc}
  arch        
  & method
  & {\tablestyle{0pt}{.9} \begin{tabular}{c} {mask} \\ {mIoU} \end{tabular}}  \\
  \shline
  ViT       & pixel-wise                                   & 20.1 \\
  ViT       & K-means                                      & 25.0 \\
  ViT       & Mean-shift~\cite{comaniciu2002mean}          & 20.7 \\
  ViT       & Spectral clustering~\cite{shi2000normalized} & 19.7 \\
  GroupViT & Ours                                         & \textbf{52.3}
  \end{tabular}
  \vspace{-.5em}
  \caption{
    \textbf{Comparisons with zero-shot baselines}. 
  }
  \label{tab:non-parametric}
  \vspace{-2em}
\end{table}

\begin{table*}[!t]
  \tablestyle{8pt}{1.1}
  \vspace{-.5em}
  \begin{tabular}{c|ccc|c|cc}
  & \multicolumn{3}{c|}{pre-training}  & & \multicolumn{2}{c}{transfer} \\ 
  arch       & model                         & dataset  & supervision           & zero-shot 
  & PASCAL VOC  
  & PASCAL Context  \\
  \shline
  ViT       & DeiT\cite{touvron2021training} & ImageNet & class       & \xmark    & 53.0 & 35.9 \\
  \hline
  ViT       & DINO\cite{caron2021emerging}   & ImageNet & self        & \xmark    & 39.1 & 20.4 \\
  ViT       & DINO\cite{caron2021emerging}   & CC12M+YFCC & self        & \xmark    & 37.6 & 22.8 \\
  ViT       & MoCo\cite{chen2021empirical}   & ImageNet & self        & \xmark    & 34.3 & 21.3 \\
  ViT       & MoCo\cite{chen2021empirical}   & CC12M+YFCC & self        & \xmark    & 36.1 & 23.0 \\
  \hline
  GroupViT & Ours                           & CC12M+YFCC & text        & \cmark    & \textbf{52.3} & \textbf{22.4}
  \end{tabular}
  \vspace{-.5em}
  \caption{
    \textbf{Comparisons with fully supervised transfer}. Zero-shot \cmark means transfer to semantic segmentation without any fine-tuning. 
    We report mIoU on the validation split of the PASCAL VOC 2012 and PASCAL Context datasets.
  }
  \label{tab:transfer}
  \vspace{-1.5em}
\end{table*}

\paragraph{Multi Stage Grouping}
In Table~\ref{tab:multi-stage}, we compare the 1-stage and 2-stage GroupViT architectures.
We also compare their visual zero-shot semantic segmentation results in Fig.~\ref{fig:vis_stage}.
We find that the 2-stage GroupViT generates smoother segmentation maps compared to its 1-stage counterpart.
To quantify the smoothness of the segmentation maps, we also report the boundary mIoU~\cite{cheng2021boundary} in Table~\ref{tab:multi-stage}, which computes the IoU of boundaries only.
The 2-stage GroupViT improves the mask mIoU of the 1-stage variant by 1.8\% and the boundary mIoU by 1.9\%.
We also train both models on a combination of the CC~\cite{changpinyo2021conceptual} and YFCC~\cite{thomee2016yfcc100m} datasets.
While the 1-stage model does not benefit as much from the expanded dataset, the 2-stage model improves significantly both in terms of the mask and boundary mIoU values by $\sim$7\%. 
These results demonstrate that our hierarchical grouping mechanism is effective, especially when training with larger datasets. 
We adopt the 2-stage GroupViT in the following experiments.

\subsection{Visualization}

\begin{figure}[]
    \centering
    \includegraphics[width=.95\linewidth]{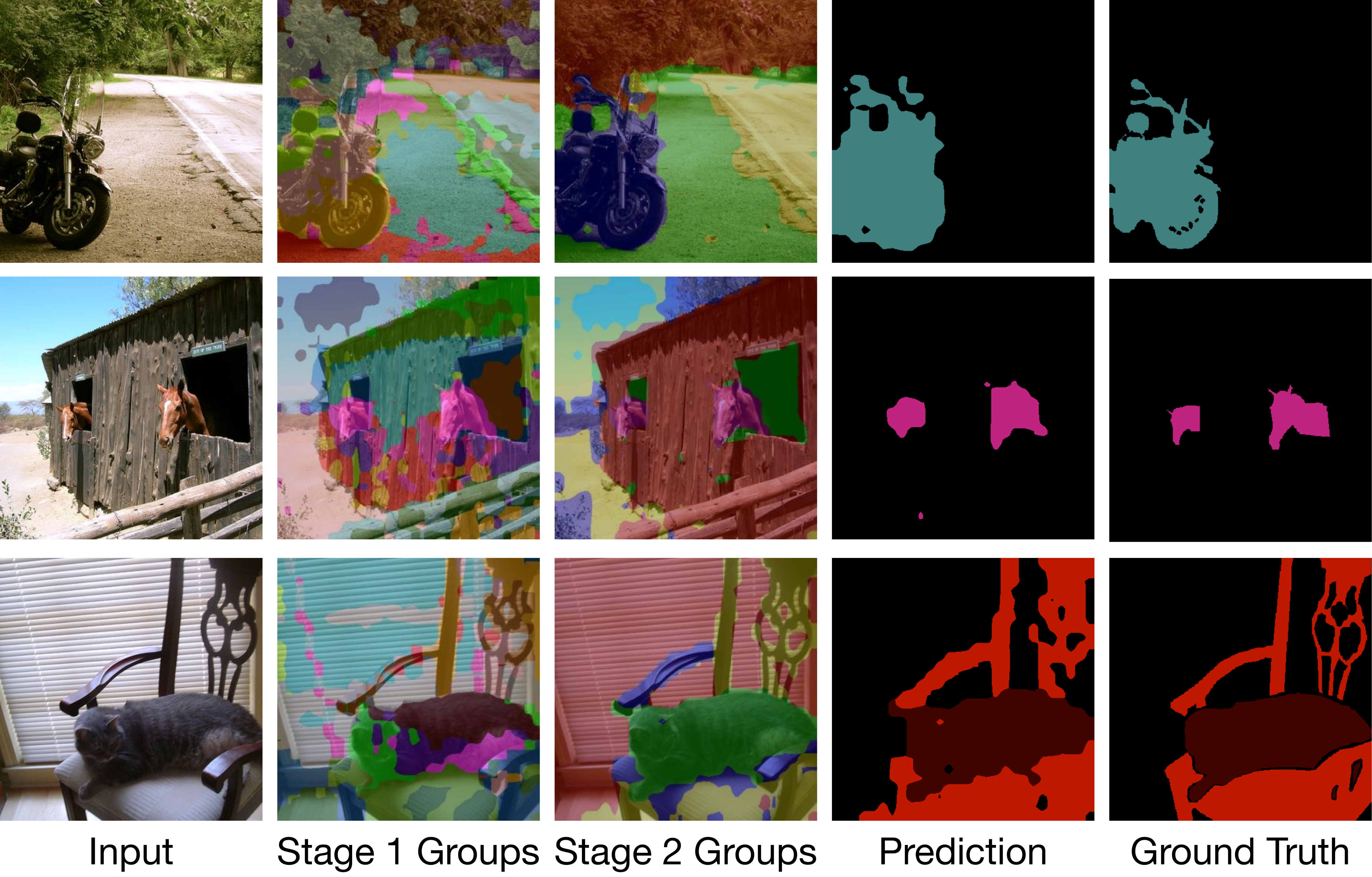}
    \vspace{-.5em}
    \caption{
    \textbf{Qualitative results on PASCAL VOC 2012.} Stage 1/2 Groups are grouping results prior to assigning labels. 
    }
    \label{fig:vis_small}
    \vspace{-0.5em}
\end{figure}

\paragraph{Qualitative Results on PASCAL VOC 2012}
We show selected qualitatively segmentation results of GroupViT in Fig.~\ref{fig:vis_small}. 
We select examples with a single object (row 1), multiple object of the same class (row 2), and multiple objects from different classes (row 3). 
GroupViT could generate plausible segmentation.
We provide more qualitative results in the supplement Sec.~\ref{sec:supp_vis}. 

\begin{figure}[]
    \centering
    \includegraphics[width=\linewidth]{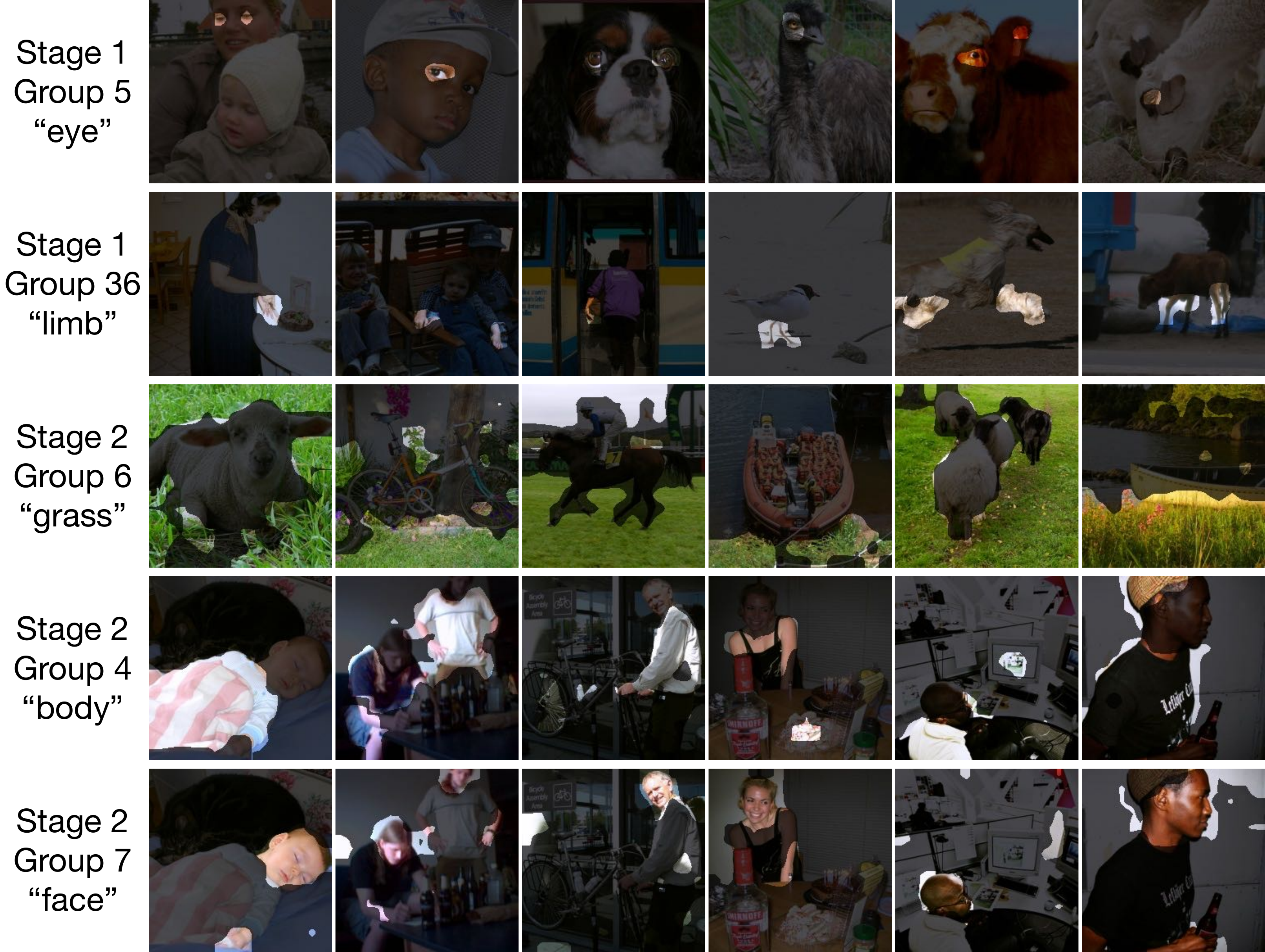}
    \vspace{-1.5em}
    \caption{
        \textbf{Concepts Learnt by Group Tokens.}. We highlight the regions that group tokens attend to in different stages. 
    }
    \label{fig:dissection}
    \vspace{-1.5em}
\end{figure}

\paragraph{Concepts Learnt by Group Tokens}
We visualize what the group tokens learn in Fig.~\ref{fig:dissection}. We select some group tokens and highlight the attention regions across images in the PASCAL VOC 2012. We found different group tokens are learning different semantic concepts.
In the first stage, group tokens usually focus on mid-level concepts such as ``eyes'' (row 1) and ``limbs''(row 2). Interestingly, the group token 36 attends to ``hands'' if people are in the image, while focusing on ``feet'' if animals like bird and dog are present. Group tokens in the second stage are more associated with high-level concepts, e.g., ``grass'', ``body'' and ``face''.
The figure also shows that the learnt concepts in the first stage could be aggregated into higher level concepts in the second stage. 

\subsection{Comparisons with Existing Methods}

We compare the zero-shot semantic segmentation performance of GroupViT with other zero-shot baselines and with methods for fully supervised transfer, based on ViT-S.

\paragraph{Comparison with Zero-Shot Baselines} 
We train ViT and a text encoder with the image-text contrastive loss defined in CLIP~\cite{radford2021learning}, for comparison.
To zero-shot transfer CLIP to semantic segmentation, during inference, we first apply non-parametric grouping on its output features. 
We then compute the similarity between the average feature of each group and the text embedding of the segmentation labels of the dataset.
In this way, any non-parametric grouping method for ViT combined with CLIP can be considered as a zero-shot semantic segmentation baseline.
We also report a ``pixel-wise" baseline, which treats each pixel as a group and performs classification independently.
As Table~\ref{tab:non-parametric} shows that GroupViT outperforms other grouping methods by a large margin.
This demonstrates that, compared to ViT trained with CLIP, our GroupViT is more effective at zero-shot transfer to semantic segmentation.
In the Table~\ref{tab:imagenet}, we also show that GroupViT's ImageNet classification performance is comparable to that of ViT.

\paragraph{Comparison with Fully-Supervised Transfer} 
We compare the performance of GroupViT with fully-supervised transfer to semantic segmentation. For fully-supervised transfer, we fine-tune a semantic segmentation head jointly with a pre-trained representation~\cite{zhao2017pyramid, chen2017deeplab} on the training sets of the PASCAL VOC 2012 and PASCAL Context datasets separately and report their performances in Table~\ref{tab:transfer}.
For a fair comparison, we employ a ViT architecture comparable to GroupViT's for all baselines. Specifically, we append a 1$\times$1 convolution layer to a pre-trained ViT, trained with images of size 224 $\times$ 224 and fine-tune the whole network with ground truth masks for 4k iterations.
During inference, we resize the input images to have a shorter side length of 448 pixels.
For fully-supervised transfer, we compare both supervised and self-supervised pre-training methods against GroupViT (Table~\ref{tab:transfer}). 
GroupViT (without fine-tuning) outperforms all variants of ViT pre-trained with self-supervision (with supervised fine-tuning) by a large margin on PASCAL VOC 2012 and is comparable to them on PASCAL Context. This implies that GroupViT, without any pixel-level annotations is able to transfer to several semantic segmentation datasets and can outperform existing state-of-the-art transfer-learning methods requiring more supervision (i.e., pixel-level labels for supervised transfer). Interestingly, on PASCAL VOC 2012, the zero-shot performance of GroupViT (mIoU of 52.3$\%$) approaches that of fully-supervised ViT (mIoU of 53$\%$) trained with both image classification and segmentation labels, which is significant.

\vspace{-.5em}
\section{Discussion}
\vspace{-.25em}
\noindent\textbf{Conclusion}
We take the first step towards learning semantic segmentation with text alone and without any explicit human supervision. 
We show that, with GroupViT, the representation learned from large-scale noisy image-text pairs can be transferred to semantic segmentation in a zero-shot manner.
This work also demonstrates that besides image classification, text supervision could also be transferred to finer-grained vision tasks, which hasn't yet been explored previously and opens up an exciting research direction.

\noindent\textbf{Limitations and Future Work}
There are two potential improvements of GroupViT to explore in the future.
Firstly, GroupViT's performance is lower on PASCAL Context versus PASCAL VOC. This happens due to the presence of background classes, e.g., \texttt{ground} and \texttt{road} in PASCAL Context, which are less likely to be labeled in text; and misclassification of correctly grouped segments into incorrect textual classes (details in supplement Sec.~\ref{sec:limit}).
Secondly, GroupViT's architecture currently doesn't integrate segmentation-specific enhancements, e.g., dilated convolutions~\cite{chen2017deeplab}, pyramid pooling~\cite{zhao2017pyramid} or a U-Net~\cite{ronneberger2015unet}.


{\footnotesize \textbf{Acknowledgements.}~Prof. Wang's lab is supported, in part, by grants from NSF CCF-2112665 (TILOS) and DARPA LwLL.}

{\small
\bibliographystyle{ieee_fullname}
\bibliography{egbib}
}

\renewcommand\thefigure{\thesection.\arabic{figure}}
\renewcommand\thetable{\thesection.\arabic{table}}
\setcounter{figure}{0} 
\setcounter{table}{0} 

\newpage
\appendix

\section{Implementation Details}

\subsection{Architecture}
The architecture of GroupViT is based on ViT-S~\cite{dosovitskiy2020image, touvron2021training} with 12 Transformer layers. 
Each layer consists of a multi-head self-attention and an MLP block. The input to each block is normalized by layer normalization~\cite{ba2016layer}.
We connect the group tokens in the different grouping stages via MLP-Mixer layers~\cite{tolstikhin2021mlp}. 
Our text-encoder consists of 12 Transformer layers, each with a hidden dimension of 256. 
Following~\cite{radford2021learning}, the Transformer operates on a lower-cased byte pair encoding (BPE) representation of the text with a vocabulary of 49,152 words.

\subsection{Fully-Supervised Transfer to Semantic Segmentation}
To implement the baselines for fully-supervised transfer to semantic segmentation, 
we fine-tune the pre-trained ViT model jointly with a 1${\times}$1 convolutional layer appended to it for pixel-wise classification.
We scale each input image by a randomly selected factor in the range of $[0.5, 2]$ and then crop random 224${\times}$224 patches from each image during training.
We use the Adam~\cite{kingma2014adam} optimizer with a weight decay of 0.05 and a learning rate 0.001. 
We train all models for 4k iterations with a batch size of 16.
During inference, we resize each input image to have a shorter side of size 448 pixels.
We open-source our code at \href{https://github.com/NVlabs/GroupViT}{https://github.com/NVlabs/GroupViT}.

\section{Qualitative Results}
\label{sec:supp_vis}
\begin{figure*}[t]
    \centering
    \includegraphics[width=.74\linewidth]{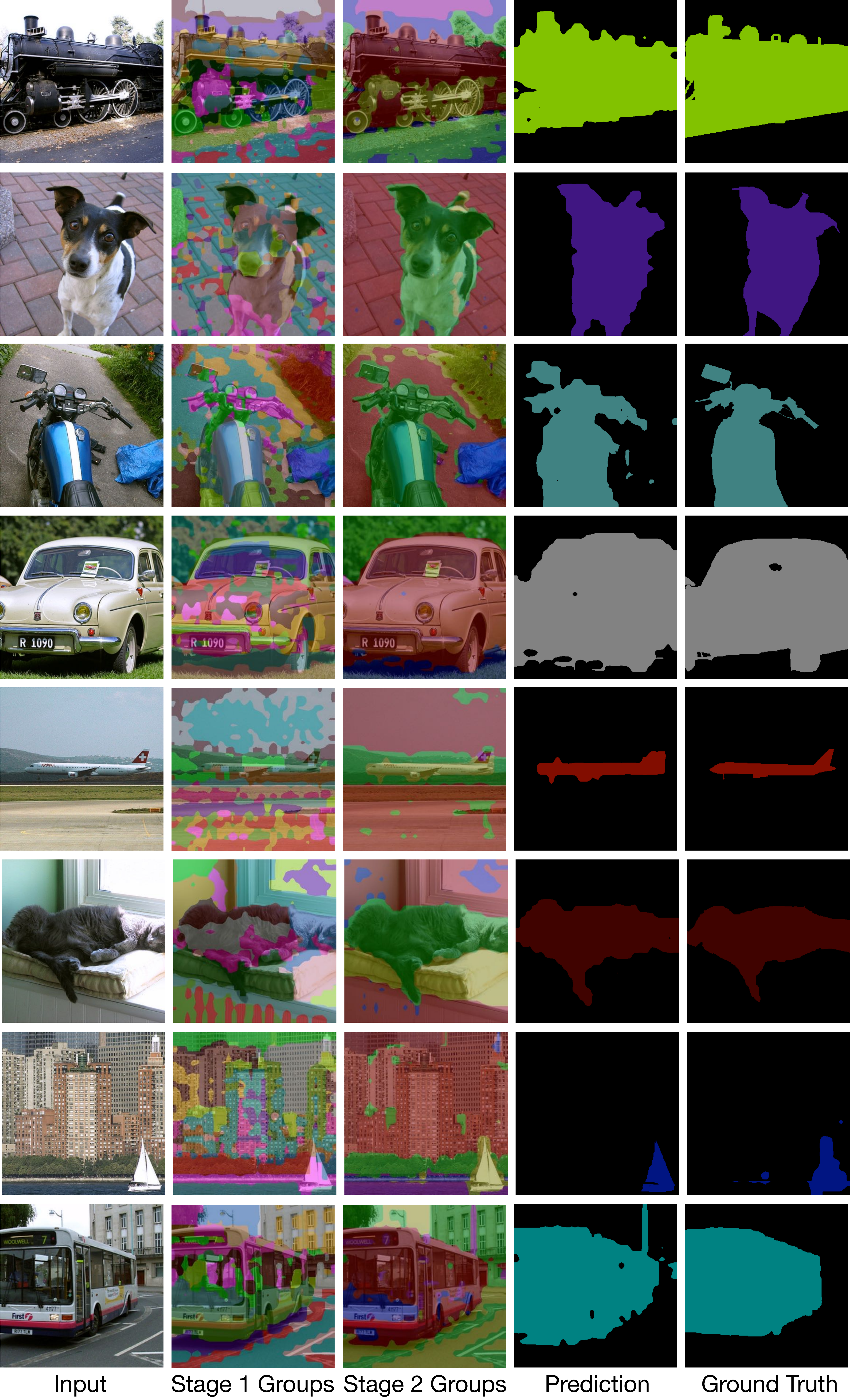}
    \vspace{-.5em}
    \caption{
        \textbf{Qualitative Results of GroupViT on PASCAL VOC 2012.} 
        The results in columns labeled ``Stage 1/2'' show grouping results prior to assigning labels, 
        where the regions belong to the same group are in the same color. 
        All these examples contain a single object from a category.
    }
    \label{fig:vis_voc1}
    \vspace{-1em}
\end{figure*}

\begin{figure*}[t]
    \centering
    \includegraphics[width=.74\linewidth]{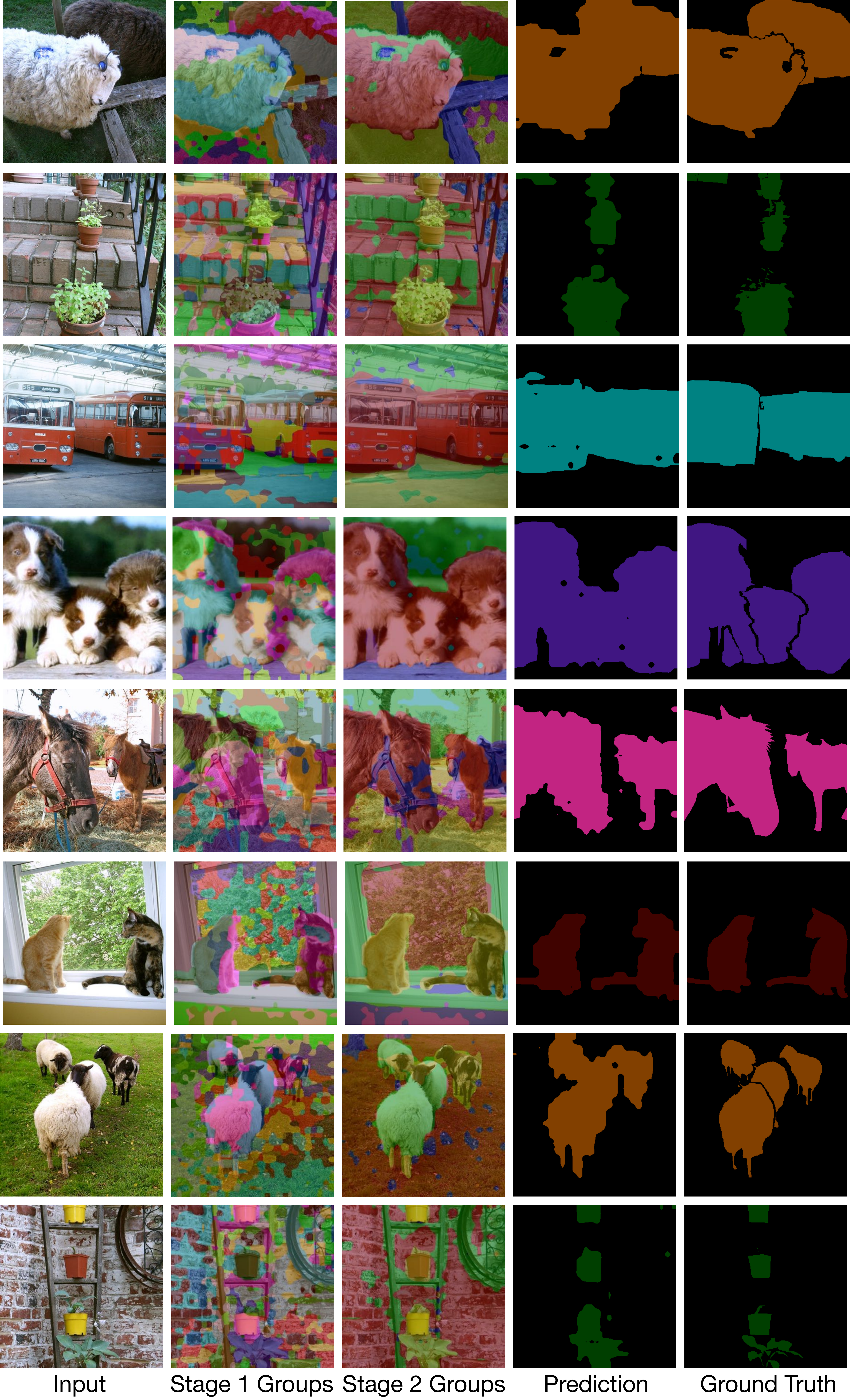}
    \vspace{-.5em}
    \caption{
        \textbf{Qualitative Results of GroupViT on PASCAL VOC 2012.} 
        The results in columns labeled ``Stage 1/2'' show grouping results prior to assigning labels. 
        The regions belong to the same group are in the same color. 
        These examples contain multiple objects from the same category.
    }
    \label{fig:vis_voc2}
    \vspace{-1em}
\end{figure*}

\begin{figure*}[t]
    \centering
    \includegraphics[width=.74\linewidth]{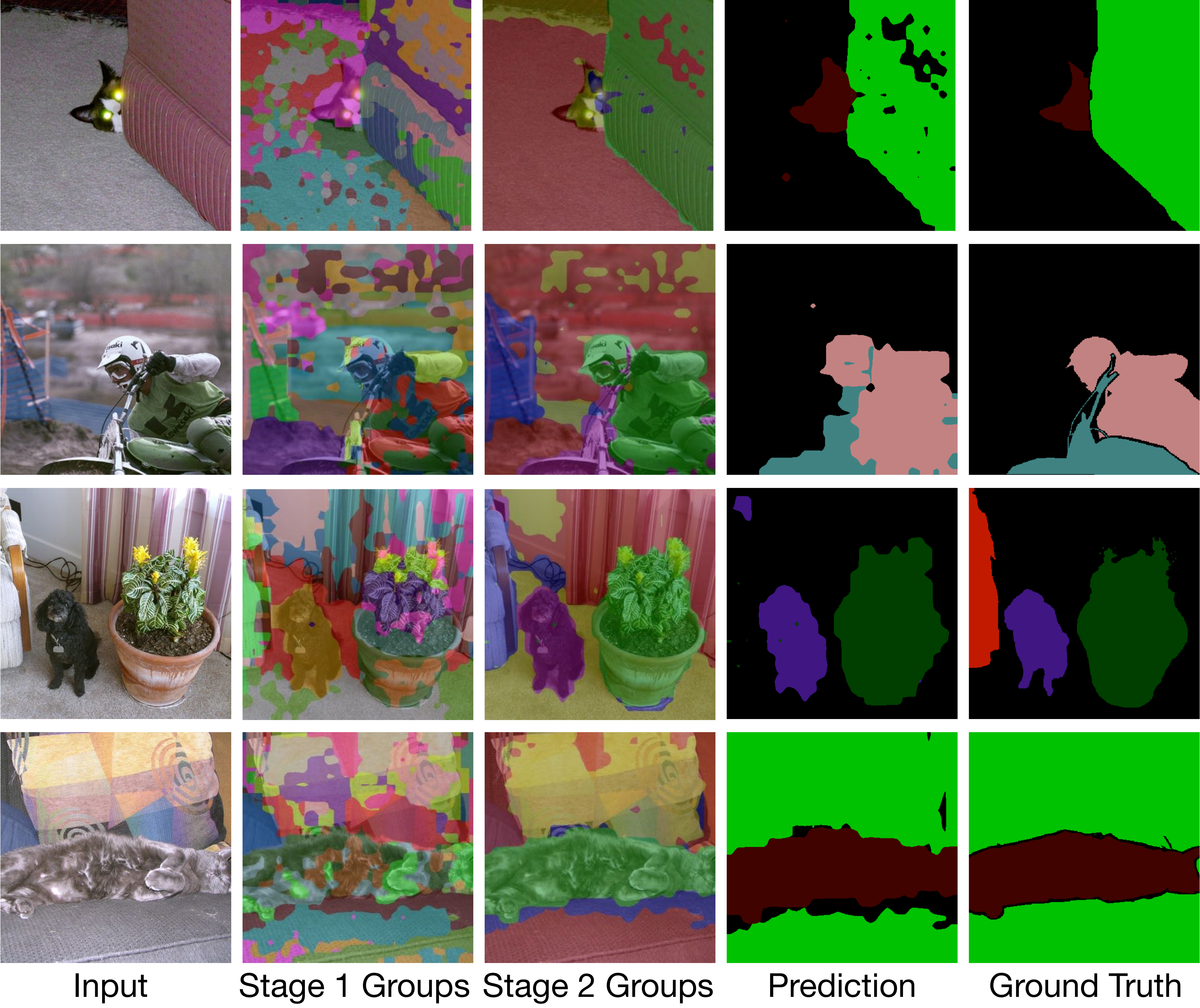}
    \vspace{-.5em}
    \caption{
        \textbf{Qualitative Results of GroupViT on PASCAL VOC 2012.} 
        The results in columns labeled ``Stage 1/2'' show grouping results prior to assigning labels, 
        where the regions belong to the same group are in the same color. 
        These examples contain multiple objects from multiple different categories.
    }
    \label{fig:vis_voc3}
    \vspace{-1em}
\end{figure*}

\paragraph{PASCAL VOC 2012}
We show additional qualitative results of GroupViT on the PASCAL VOC 2012 dataset, i.e. examples with single object in Fig.~\ref{fig:vis_voc1}; multiple objects from the same category in Fig.~\ref{fig:vis_voc2}; and multiple objects from different categories in Fig.~\ref{fig:vis_voc3}. Observe that GroupViT successfully groups and correctly classifies the objects in these various challenging scenarios.

\begin{figure*}[t]
    \centering
    \includegraphics[width=.74\linewidth]{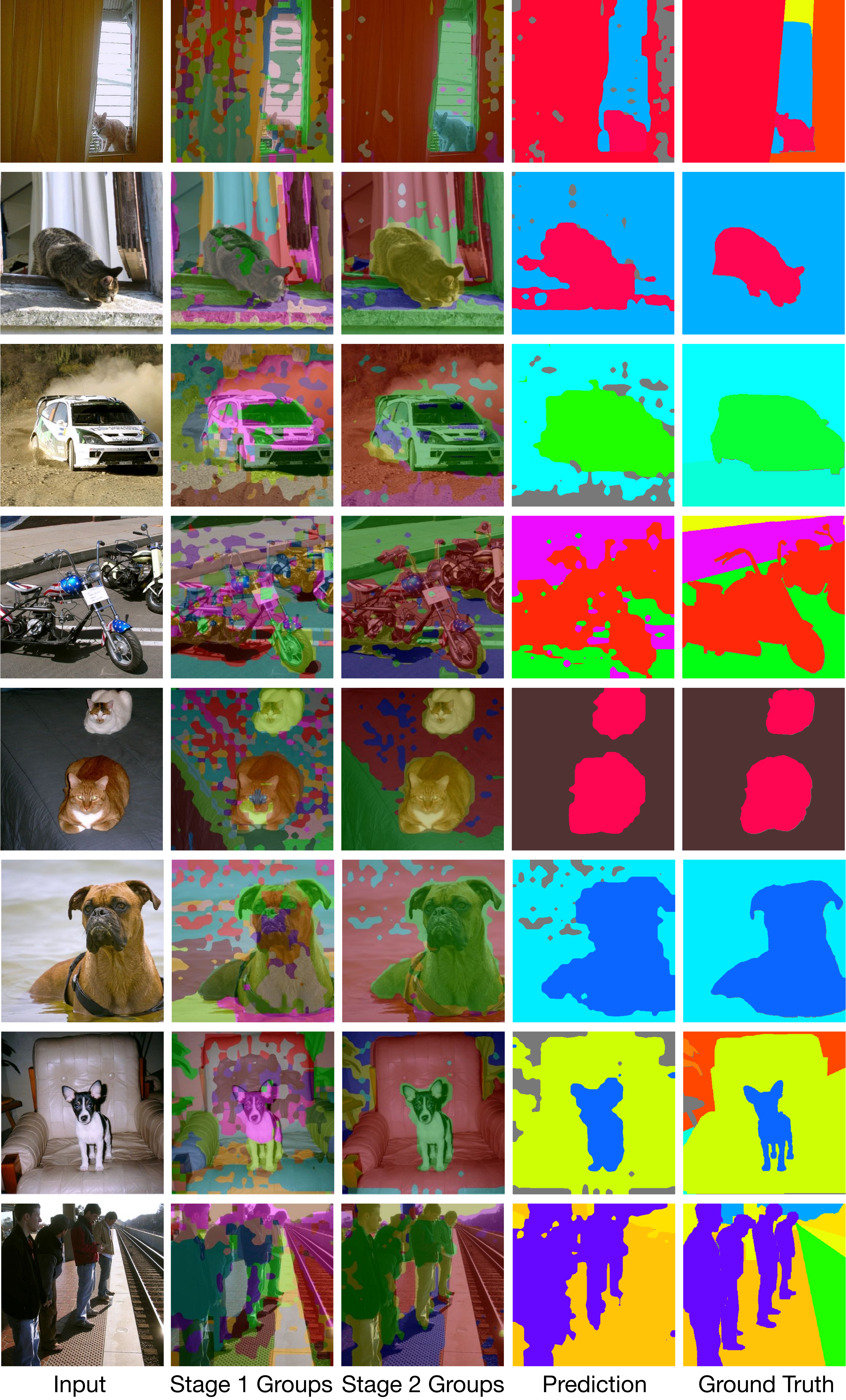}
    \vspace{-.8em}
    \caption{
        \textbf{Qualitative Results of GroupViT on PASCAL Context.} 
        Columns labeled ``Stage 1/2" show grouping results prior to assigning labels, 
        where the regions belong to the same group are in the same color. 
        GroupViT can successfully segment \textit{object} and \textit{stuff} classes, e.g. \texttt{cat} and \texttt{window} in row 2, \texttt{dog} and \texttt{water} in row 6.
    }
    \label{fig:vis_ctx}
    \vspace{-1.5em}
\end{figure*}

\paragraph{PASCAL Context}
We show more qualitative results of GroupViT on the PASCAL Context dataset in Fig.~\ref{fig:vis_ctx}. The PASCAL Context dataset annotates not only \textit{object} classes from PASCAL VOC 2012, e.g. \texttt{car} and \texttt{dog}, but also \textit{stuff} classes related to the context, e.g. \texttt{sky} and \texttt{water}.
Observe that GroupViT successfully segments \textit{object} and \textit{stuff} classes in the PASCAL Context dataset, e.g., \texttt{cat} and \texttt{window} in the second row, and \texttt{dog} and \texttt{water} in the sixth row.

\section{Additional Experiments and Analysis}

\subsection{Image Classification}
We compare the performance of the GroupViT and ViT architectures for the task of object classification on ImageNet. 
Following CLIP~\cite{radford2021learning}, here we train both architectures using supervision from text only via an image-text contrastive loss. 
In Table~\ref{tab:imagenet}, we report both the zero-shot and the linear probing accuracy on the ImageNet~\cite{deng2009imagenet} validation split. The zero-shot and linear probing evaluation follow the same setting as CLIP~\cite{radford2021learning}.
GroupViT's ImageNet classification performance is comparable to (if not better than) that of ViT, thus demonstrating that our proposed grouping mechanism enhances the baseline ViT architecture with the capability to perform semantic pixel grouping and zero-shot transfer to semantic segmentation, without affecting its object classification performance.

\begin{table}[h]
  \tablestyle{6pt}{1.1}
  \vspace{-1em}
  \begin{tabular}{c|cc}
  model    
  & zero-shot Acc@1 & linear Acc@1 \\
  \shline
  ViT       & 42.4      & 69.2   \\
  GroupViT & 42.9      & 69.8   \\
  \end{tabular}
  \vspace{-.5em}
  \caption{
    \textbf{ImageNet Accuracy}.
  }
  \vspace{-1em}
  \label{tab:imagenet}
\end{table}

\subsection{Mask Probing}
We follow the procedure outlined in DINO~\cite{caron2021emerging} to evaluate the quality of the masks generated by GroupViT and by the baseline ViT model pre-trained using prior methods in a fully supervised~\cite{touvron2021training}, self-supervised~\cite{chen2021empirical,caron2021emerging} or text-supervised~\cite{radford2021learning} manner.
For the ViT models, similar to DINO~\cite{caron2021emerging} for each final attention head, we compute its similarity to the \texttt{[CLS]} token and derive an attention mask for the pixels with the highest attention values. We then compute the Jaccard similarity of each head's attention mask to the ground truth mask and retain the attention mask with the highest similarity.
As for GroupViT, it does not have a multi-head design in the Grouping Block. Thus, we directly select the group most similar, as measured by the Jaccard index, to the ground truth mask for each image. 
As Table~\ref{tab:mask_probing} shows, the mask probing result of GroupViT is significantly better than that of all variants of the baseline ViT architecture. 
Hence, compared to ViT, our GroupViT more effectively groups semantically-related visual inputs together.

\begin{table}[]
  \tablestyle{2pt}{1.1}
  \vspace{-1em}
  \begin{tabular}{c|ccc|c}
  arch        & model                                           & dataset  & supervision 
  & {\tablestyle{0pt}{.9} \begin{tabular}{c} {Jaccard} \\ {Similarity} \end{tabular}}  \\
  \shline
  ViT       & Random                         & -      & -           & 23.6              \\
  ViT       & DeiT\cite{touvron2021training} & ImageNet & class       & 24.6              \\
  ViT       & MoCo\cite{chen2021empirical}   & ImageNet & self        & 28.2              \\
  ViT       & DINO\cite{caron2021emerging}   & ImageNet & self        & 45.9              \\
  ViT       & DINO\cite{caron2021emerging}   & CC12M+YFCC & self        & 41.8              \\
  ViT       & CLIP\cite{radford2021learning}   & CC12M+YFCC & text        & 28.6              \\
  GroupViT & Ours                           & CC12M+YFCC & text        & 51.8              
  \end{tabular}
  \vspace{-.5em}
  \caption{
    \textbf{Comparison of mask probing performance} GroupViT outperforms all other variants of the baseline ViT architecture at effectively grouping image regions on semantic groups. 
  }
  \vspace{-1em}
  \label{tab:mask_probing}
\end{table}

\begin{figure*}[h]
    \centering
    \includegraphics[width=.74\linewidth]{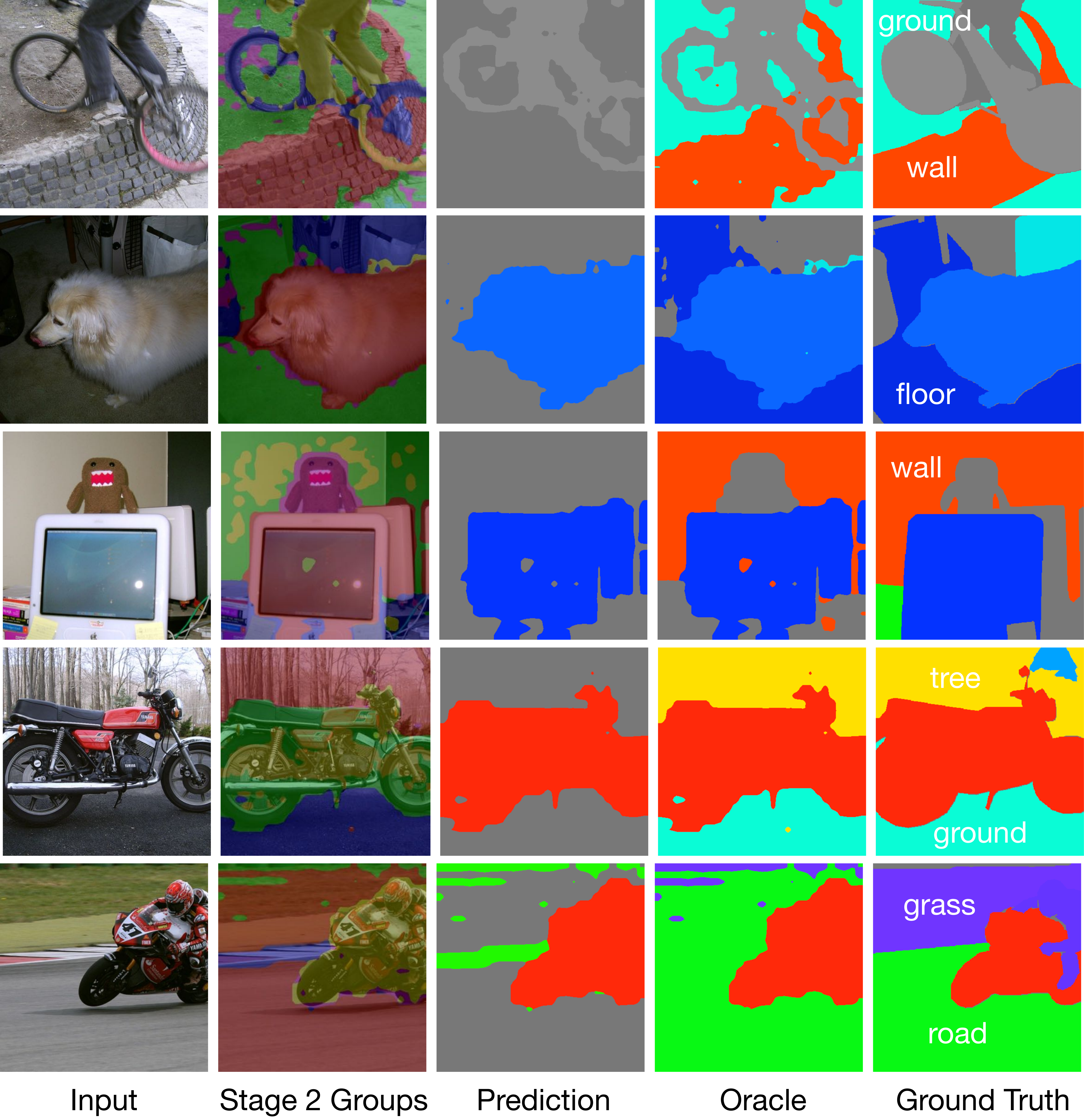}
    \vspace{-.5em}
    \caption{
        \textbf{Failure cases on PASCAL Context.} ``Oracle" shows the results of assigning groups to segmentation classes based on their IoU with the ground truth masks. Although GroupViT successfully groups \textit{stuff} classes, e.g. \texttt{ground}, \texttt{road} and \texttt{wall}, it is not able to classify them correctly using the similarity between the visual and text embedding.  
    }
    \vspace{-1em}
    \label{fig:vis_ctx_fail}
\end{figure*}

\subsection{Limitations}
\label{sec:limit}

We found that the mIoU of GroupViT on PASCAL Context is significantly lower than that on PASCAL VOC 2012. 
This could be attributed to the presence of background classes in PASCAL Context,  
e.g., \texttt{ground}, \texttt{road} and \texttt{wall} that result in low IoU ($\sim$1.5) on zero-shot transferring GroupViT to semantic segmentation on PASCAL Context. 
Through visual inspection, we found that while the pixels belonging to these background classes are typically correctly grouped into a single group by GroupViT, the group as a whole may be miss-classified into the wrong class on being compared to the text embedding of the various class labels. We hypothesize that this, in turn, happens due to the low probability of the background classes being described in textual sentences used during training. We show examples of such failure case in Fig.~\ref{fig:vis_ctx_fail}.
We further conduct an oracle experiment to verify this finding. 
In the oracle experiment, for each output group from GroupViT, we compute its IoU with all ground truth masks and assign to each group the class label that results in the the maximum IoU. 
This represents the upper bound of GroupViT's performance since here we leverage ground truth masks to predict each group's class label. 
We use our 2-stage GroupViT trained on CC12M and YFCC datasets for this oracle experiment, which is the same model labeled "Ours" in Table~{\textcolor{red} 5} of the main paper.
We report the oracle experiment's results on PASCAL Context in Table~\ref{tab:oracle_ctx}.
The large gap between the performance of the original and oracle mIoU values on the PASCAL Context dataset, shows that while GroupViT's grouping results are reasonably good, there is room to further improve the groups' classification to segmentation class labels via image-text embedding similarity.

\begin{table}[!h]
  \tablestyle{6pt}{1.1}
  \vspace{-1em}
  \begin{tabular}{c|c|c|c}
  arch      
  & dataset
  & mask mIoU
  & {\tablestyle{0pt}{.9} \begin{tabular}{c} {oracle} \\ {mask mIoU} \end{tabular}}  \\
  \shline
  GroupViT & PASCAL VOC& 52.3      & 73.7 \\
  GroupViT & PASCAL Context& 22.4      & 54.6 \\
  GroupViT & COCO& 24.3      & 54.0 \\
  \end{tabular}
  \vspace{-.5em}
  \caption{
    \textbf{Original versus oracle results}. 
  }
  \vspace{-1em}
  \label{tab:oracle_ctx}
\end{table}

\subsection{COCO Dataset}
We evaluate the performance of GroupViT on the COCO dataset~\cite{lin2014microsoft}, which contains 80 object classes. We combine the instance masks of the same category to get the semantic segmentation mask for each image.
We  report semantic segmentation mIoU on COCO in Table~\ref{tab:oracle_ctx}.
It demonstrates that GroupViT is able to transfer to complex datasets with various number of classes.

\subsection{Training on RedCaps}
To show that our approach is generalizable to other training datasets, besides CC~\cite{sharma2018conceptual, changpinyo2021conceptual} and filtered YFCC~\cite{thomee2016yfcc100m}, we also train GroupViT on the recently released RedCaps dataset~\cite{desai2021redcaps}, which contains 12 millions image-text pairs from Reddit, of similar size as filtered YFCC.
We report mIoU for zero-shot transfer to various image segmentation benchmarks datasets in Table~\ref{tab:redcap}.
Replacing YFCC with RedCaps yields similar accuracy on Pascal VOC, Pascal Context and COCO datasets.
It demonstrates that GroupViT is able to learn grouping with properly filtered image text pairs.

\begin{table}[!h]
  \tablestyle{5pt}{1.1}
  \vspace{-1em}
  \begin{tabular}{c|c|c|c|c}
  arch      
  & {\tablestyle{0pt}{.9} \begin{tabular}{c} {Training} \\ {Dataset} \end{tabular}} 
  & {\tablestyle{0pt}{.9} \begin{tabular}{c} {PASCAL} \\ {VOC} \end{tabular}} 
  & {\tablestyle{0pt}{.9} \begin{tabular}{c} {PASCAL} \\ {Context} \end{tabular}} 
  & COCO \\
  \shline
  GroupViT & CC+YFCC & 52.3      & 22.4 & 24.3 \\
  GroupViT & CC+RedCaps & 50.8      & 23.7 & 27.5 \\
  \end{tabular}
  \vspace{-.5em}
  \caption{
    \textbf{Results trained with CC+RedCaps}. 
  }
  \vspace{-1em}
  \label{tab:redcap}
\end{table}

\end{document}